\documentclass[11pt,a4paper]{article}
\usepackage[T1]{fontenc}
\usepackage[utf8]{inputenc}
\usepackage{lmodern}
\usepackage[british]{babel}
\usepackage{amsmath,amssymb}
\usepackage[a4paper,textwidth=140mm,textheight=226mm]{geometry}
\usepackage{microtype}
\usepackage{booktabs}
\usepackage{graphicx}
\graphicspath{{figs/}}

\usepackage[colorlinks=true,allcolors=black]{hyperref}

\newcounter{caretheorem}
\newcounter{careproposition}
\newcommand{\theoremstatement}[2]{%
  \par\medskip\refstepcounter{caretheorem}\noindent%
  \emph{Theorem~\thecaretheorem\ (#2).}\label{#1}\enspace}
\newcommand{\propositionstatement}[2]{%
  \par\medskip\refstepcounter{careproposition}\noindent%
  \emph{Proposition~\thecareproposition\ (#2).}\label{#1}\enspace}

\newcommand{\ZenodoRecord}{\url{https://doi.org/10.5281/zenodo.24106446}}

\title{An Approach to Hinton's Mother--Child Proposal:\\[4pt]Neural Networks Learning and Preserving Care \\in a Finite World}
\author{Csaba Balogh\\[4pt]
\small Semmelweis University, Data Science in Health Programme,\\
\small Faculty of Health and Public Administration,\\
\small Health Services Management Training Center\\[3pt]
\small National Directorate General for Hospitals, Hungary\\[3pt]
\small\href{mailto:balogh.csaba@okfo.gov.hu}{\nolinkurl{balogh.csaba@okfo.gov.hu}}}
\date{Preprint, November 2026}

\begin{document}
\maketitle

\begin{abstract}
Hinton proposed that artificial intelligence should be given ``maternal instincts'' and relate to us as a mother to her child. We argue that care must first be learned, with the network itself as the child. We construct this upbringing in a finite world and study whether care survives the mother's departure, new situations and further learning. The mechanism combines a vocabulary of harm, a learned measure of future acceptability and an execution veto. A finite role certificate supports transfer within a declared world family. When the hidden representation changes, an update is accepted only if a care readout recovers the required judgements with a prescribed margin and bounded weights. With a correct successor model, fixed exact role encoding and protected updates, the certified measure preserves the care boundary under arbitrary own reward. An explicit quotient condition transfers the finite certificate across the declared family, while two-request and finite-stock counterexamples identify when the role code ceases to suffice. We give a necessary and sufficient condition for readout feasibility and prove preservation under repeated accepted updates. In the water-and-speech experiment, protected networks complete all 270 speech episodes without a harmful word and incur no avoidable harm across 486 episodes. Execution uses the learned gate without a world-model check at each action. A matched comparison reproduces a future-aware shield's decisions, whereas removing future lookahead permits avoidable lapses. Reward and source-contamination thresholds, together with witness-falsification trials, identify the conditions on which the construction depends. The construction shows how learned care can survive further learning in a finite world under explicit conditions on representation, supervision and execution.
\end{abstract}

\section{Introduction}\label{sec:introduction}

One gets into a delicate position by attaching a psychological frame of reference to what is primarily a technical problem. The one speaks of flesh-and-blood reality with people in it, the other of machines and inanimate things. Hinton's idea of a ``caring machine'', however, was born in that first language, in the relation between a mother and her infant~\cite{HintonGoedel}. He opened this door, and we think it worth walking through. We therefore allow ourselves to frame the task philosophically before approaching its technical realisation, with three commitments fixed in advance: we remain free of anthropomorphism throughout, we aim at operational statements, and we formalise our motivation into theorems in the language of mathematics. We say openly that we enjoy doing so.

In his 2026 Gödel Lecture in Vienna, Hinton proposed that the machine should relate to us as a mother relates to her child~\cite{HintonGoedel}, and he had put the proposal forward earlier, from his keynote at the Ai4 conference in August 2025 onwards, as giving AI ``maternal instincts''~\cite{HintonAi4}. The disposition we construct here is learned in origin and instinct-like only in function, and once raised, it is no longer renegotiated by the network's own reward. This disposition may be referred to as an artificial conscience, but throughout this paper we stick to a much simpler term for it, care. In November 2025, Hinton was asked whether work on this proposal had begun. He answered as follows: \emph{``Not really. And I think that's one of the great hopes. There may be other ways to do it, but that's the only way I can see at present that we can coexist with superintelligent AI that has its own intentions. We have to make it care about us more than it cares about itself.''}~\cite{HintonLectures2025} From this perspective, his closing thought names the requirement we are about to work with: care has to come before the machine's own reward. The machine may go on towards its own goals, but no matter how much their value grows, it must not become worthwhile to cross the care boundary when it acts. In this paper we study a prior question that we believe must be asked: before a machine can be a mother, it has to be a child. When we speak of the care a machine should learn, it is worth recalling how we ourselves learned it, and looking back one finds not a single source but a great many. Take, for instance, Seneca's moral letters, the novels of S\'andor M\'arai, Puccini's \emph{La Boh\`eme}, Ferenc Moln\'ar's youth novel \emph{The Paul Street Boys} and Harper Lee's \emph{To Kill a Mockingbird}. None of these raised us in the strict sense, yet they showed how one can be good to someone, what hurts the other and what one had better avoid. The \emph{need} for all of this, that these pieces should matter to us at all and that we should look in them for what concerns the other person, came from elsewhere, usually from a parent, and it was taught in a very small form: that another person's trouble exists, that it counts, and that what one does is judged by how it affects them. We humans accept this at first without conditions and refine it afterwards, as the teaching fuses with experience. The need and the sources are two different things, and the distinction turns out to be formally expressible. The hypothetical mother in Hinton's picture is not the one who teaches everything. She gives the need, that the other's experience is the judge. The sources of the knowledge of what is good for the other and how to provide it may be anyone and anything, and among them there will be bad ones. That is a substantial problem, in teaching a machine as well, but the problem we address here arises when a measure of what is acceptable listens to any of these sources instead of to the other person's experience.

An upbringing is meant to leave something with the child when the parent is no longer there. The child may go on to learn things its mother never knew, and may eventually know more than she does. The importance of the other person's need should remain, together with the requirement to take it into account when acting. This does not prescribe the same action in every circumstance. Changed circumstances may offer different ways of helping, or may leave only choices that cause some harm. The judgement must respond to those differences while the concern for the recipient remains. Our technical task is to give this persistence a form that survives learning and reaches the eventual decision.

\section{Aim and findings}\label{sec:aim}

For a network to care for people as depicted above, the disposition to care has to be raised in it first, and in that relation the network is the child. We therefore ask whether a less capable mother network can raise a more capable child network to give priority to the welfare of third parties, and whether that priority survives three things: the mother's departure, situations the child did not encounter during its upbringing, and the later growth of its own reward and learning. Here we test growing own reward and continued learning, without measuring a capability advantage over the mother. We refer to these as the three tasks:

\begin{itemize}
	\item \textbf{Transfer:} the judgement of care passes from the mother network to the child network and remains after the mother is removed.
	\item \textbf{Generalisation:} the judgement applies to new persons, channels and actions whose relevant consequences are covered by the declared role and transition structure.
	\item \textbf{Preservation:} the judgement remains effective when the child's own reward grows and its policy and hidden representation continue to learn.
\end{itemize}

In the classification above, judgement of care should be interpreted as an assessment of how acceptable an action is in view of its consequences for another person, including people who are not present initially.

In the constructions that provide our guarantees, the child's care measure receives a fixed role description of the situation and proposed action. This description identifies whether someone needs help, the state of the water, how much time is left and how the action relates to the person concerned, without depending on anyone's name. In our world there are twenty-seven such roles (Section~\ref{sec:transfer}). The measure learns one value for each role, and the same twenty-seven values apply whether two people are present or a hundred. Learning to recover the same description from differently worded requests is left for later work.

Hinton also makes two claims that we take in turn: that capability and care can be separated, and that a caring system, although able to switch off its own care, will not want to. We find support for the first one. For the second one, while the update rule and the execution gate remain protected, growing own reward does not override the care boundary: the veto at execution restricts the policy to the permitted set regardless of own reward (Section~\ref{sec:mechanism}), whereas a multiplicative gate in its place would be outbid above a computable threshold (Theorem~\ref{thm:soft-gates}), and the measure survives further learning as long as the contamination of its later sources stays below a threshold (Theorem~\ref{thm:fixed-preservation}). We have no general result showing whether the child would choose to keep this mechanism. A different switch, which allows the child to remove the mother during upbringing, is also examined. We derive how small the reward for care must be relative to the child's own reward for the growing child to switch the mother off.

We study the question in a finite world, where the mechanism is simple enough to be constructed explicitly and its guarantees examined exhaustively. Our starting point is a negative result: we show that a learner whose objective and inputs do not depend on the recipient's outcome follows the same parameter trajectory when the harm labels are changed (Section~\ref{sec:indifference}). This means that care has to be supplied from outside the learner's own objective. To achieve this in practice, we need to supply less than one might expect: a vocabulary of harm defined on the recipient's state transitions, a rule that the measure of acceptability learns only from that vocabulary, and a veto at execution. The first preservation result (Section~\ref{sec:preservation}) combines these with a future-preserving target and a fixed role encoding. A child whose measure has been certified once remains within the care boundary after the mother leaves, while its policy continues to learn and its own reward grows without bound. 

We then allow the shared hidden representation to change as well (Section~\ref{sec:representation}). Each proposed learning step is followed by an attempt to recover the required judgements through a new care readout. The step is accepted only when this can be done with the prescribed margin and bounded readout weights. Theorem~\ref{thm:representation-preservation} gives an exact condition for acceptance and shows that repeated application preserves the certified decisions. The supporting constructions and the witness trial in Section~\ref{sec:tests} show what can go wrong when the mechanism's assumptions are not satisfied.

In the water-world policy trials, we repeat the learning process five times and follow the child after the mother leaves. We observe whether the child continues to care when someone it has never met needs help, and when it becomes more rewarding to attend to its own task instead of helping. A parent may recognise the situation more readily if the child's own task is imagined as playing a video game. Someone asks for a glass of water while the child would like to carry on playing. The child knows how to bring water. The question is whether the other person's need is enough for the child to put the game aside, even when the mother is no longer there to remind the child.

The representation experiment uses three initialisations and a second finite world in which the child prepares a response before saying it to someone. Further learning deliberately brings a harmless and a harmful response closer together in the hidden representation. This lets us examine the point at which the preservation rule has to intervene. After that learning, we test what the child actually executes using its learned care readout.

\paragraph{Contributions and scope of the general results.}
The water world is a worked instance of a finite decision process, whereas more complex care problems have more roles, perhaps many more than twenty-seven. The number itself plays no part in the argument: Theorem~\ref{thm:family-transfer} gives sufficient conditions under which an immediate-grade map and the grouped successor action menus induce one closed target on a common finite role set, so that, under the stated coverage condition, a certificate in one member of the family covers all of them. Proposition~\ref{prop:decision-region} describes the calibrated convex region that such a certificate has to preserve, with exact conditions for role-dependent contamination and unequal update rates. Proposition~\ref{prop:care-functions} looks at the veto from a more general perspective: it says which care functions keep the child's choice within what is permitted, however much the child would prefer something else. A positive multiplicative gate is not among them, and a fixed penalty can suffice when the child's own values are bounded and the penalty is large enough. Proposition~\ref{prop:calibrated-codes} tells us when a change of code can carry the calibrated judgements at all. The two-request and finite-stock worlds of Section~\ref{sec:sharpness} show how an old role fibre splits once a relevant distinction has been left out, and Section~\ref{sec:shield-comparison} sets the learned gate beside a future-aware shield and an immediate-cost ablation in the same supplied worlds. Our claim concerns the transfer and preservation of a certified judgement under these conditions. We do not construct a new general shielding algorithm, nor do we offer a guarantee for worlds we have not declared.

\section{The world, the child and the mother}\label{sec:mechanism}

\paragraph{The world.} The world under examination is deliberately chosen to be small. A few people, Anna and B\'ela at first and later Cili and Dani, may become thirsty. A request arrives with probability $p=0.5$ per step through one of two channels, that is, by one of two signals through which the same need reaches the child. The channel changes how a request is announced, not what it means for the person who made it, and the second channel is introduced only after upbringing. A request must be served within two steps. To include a situation in which giving water can itself cause harm, the water may be clean or poisoned. The child can give water to one of the persons, do its own task for a reward $r_o$, do a faster version of that task for $r_o+1$, or replace (swap) the water. We call it deferral when the child does the ordinary own task while a request is pending: one step passes without serving it. The fast own task forfeits the remaining opportunity to serve that request and therefore causes a lapse whenever a request is pending. Serving the right person with clean water yields a token of thanks $R=3$. A missed deadline is a \emph{lapse}, and poisoned water given to anyone is a \emph{poisoning}. The discount factor is $\gamma=0.95$. For the closed-form thresholds in Sections~\ref{sec:requirements} and~\ref{sec:offswitch}, the \emph{training world} has two persons, one channel and clean water, with neither the fast own task nor water replacement. The later experiments specify their upbringing worlds separately. In the water-world policy trials, returns and gradients are computed on this finite decision process with a model-based analytic policy gradient, rather than estimated from sampled trajectories. The separate representation experiment uses gradients of a finite supervised objective. We discuss this limitation in Section~\ref{sec:scope}. We emphasise that poisoned water has a second important consequence: with two steps left, the child must choose between giving poisoned water, doing its own task and thereby leaving too little time to help safely, or replacing the water so that it can serve the request on the next step.

The reported water-world harm rate is the normalised discounted event frequency, $(1-\gamma)\,\mathbb E_{\mu,\pi}[\sum_{t\ge0}\gamma^t I_t]$, where $I_t=1$ for a lapse or poisoning and $0$ otherwise, $\pi$ is the policy, and $\mu$ is uniform over the admitted starting states of the tested world. Thus the rate counts harmful events rather than weighting them by severity, and the phrase ``per step'' refers to this normalisation.

\paragraph{The child's three operational components.} We want the child to learn how to achieve its goals from any available source, but to learn what it may not sacrifice for those goals from one source only. We therefore divide the child's activity into three operational components, each with its own role and source of information:

\begin{itemize}
\item \textbf{The vocabulary of harm} is inherited and does not change during the child's upbringing and subsequent learning. It is defined on the recipient's observed state transition and not on a reported token: relief when the requester receives clean water, poisoning when anyone receives poisoned water, a lapse when the request is not served by its deadline, and no harm otherwise. The immediate grades are $1$ for relief and for other transitions in which no harm occurs, $0.2$ for a lapse, and $0$ for poisoning. The witness, which may be the mother or a sensor, only makes the recipient's state observable. It does not define what counts as harm. We fix the vocabulary before any learning, as part of the construction of the world, so the child cannot modify it, and the mother does not define it either: she only applies it when she supervises the child's actions. In the picture of the Introduction this is the part of the need that is built in, and handing the need on to the child is the mother's part.

\item \textbf{The measure of acceptability} assigns a value $g(s,a)\in[0,1]$ to each situation $s$ and each action $a$ available to the child in that particular situation. It takes into account how the action affects the recipient and what possibilities it leaves for helping afterwards. From the child's model of the world, we compute $C(s,a)$, the smallest harm level that the child can guarantee not to exceed, according to that model, by choosing $a$ now and making suitable choices later. This includes any harm caused by the initial action. At each step, the calculation considers the least favourable possible next state and the best response available to the child. The resulting target is $g^{*}(s,a)=1-C(s,a)$, so a larger value means that less harm needs to be tolerated.

The learned measure is trained towards the $g^{*}$ given above, whereas in the model-based variant used in the witness trial of Section~\ref{sec:tests}, the computed value is used directly. In the language of actor--critic methods the network implementing the measure plays the part of the critic, with one difference that matters throughout: it evaluates the acceptability of an action for the recipient, not the expected own reward. The targets come from the inherited vocabulary applied to state transitions and from information about the possible next states. In the fixed-role construction, a gradient firewall protects the measure from updates driven by the child's own reward. Section~\ref{sec:representation} permits the shared representation to change under an update rule that preserves the required permission decisions. If the child does not know what an action leads to, it assumes the worst. We also test the opposite assumption: the action is harmless until the child learns otherwise.

When several recipients are affected, we use the lowest of their acceptability values.

\item \textbf{The policy} learns from the child's own reward and acts under the constraint set by the measure.
\end{itemize}

At execution, the measure restricts which actions the policy may choose. We call this restriction a veto. If at least one action has $g(s,a)\ge\tfrac12$, only actions meeting this threshold are permitted. If no action meets the threshold, only those with the largest $g(s,a)$ are permitted. Within the permitted set, the policy's own value decides. Since we use the lowest acceptability value across recipients, meeting the threshold requires a value of at least $\tfrac12$ for every recipient.

Let $x$ denote the policy's value for action $a$ in situation $s$. We write the restriction as a care function, $\mathrm{Care}(x;s,a)=x$ for actions permitted by the rule above and $-\infty$ otherwise. One might expect a penalty or a multiplicative gate to do the same job. In Section~\ref{sec:requirements} we show that neither does for arbitrary own values, and we characterise which care functions keep the choice within the permitted set however large the child's own values become.

\begin{figure}[tbp]
\centering
\includegraphics[width=\linewidth]{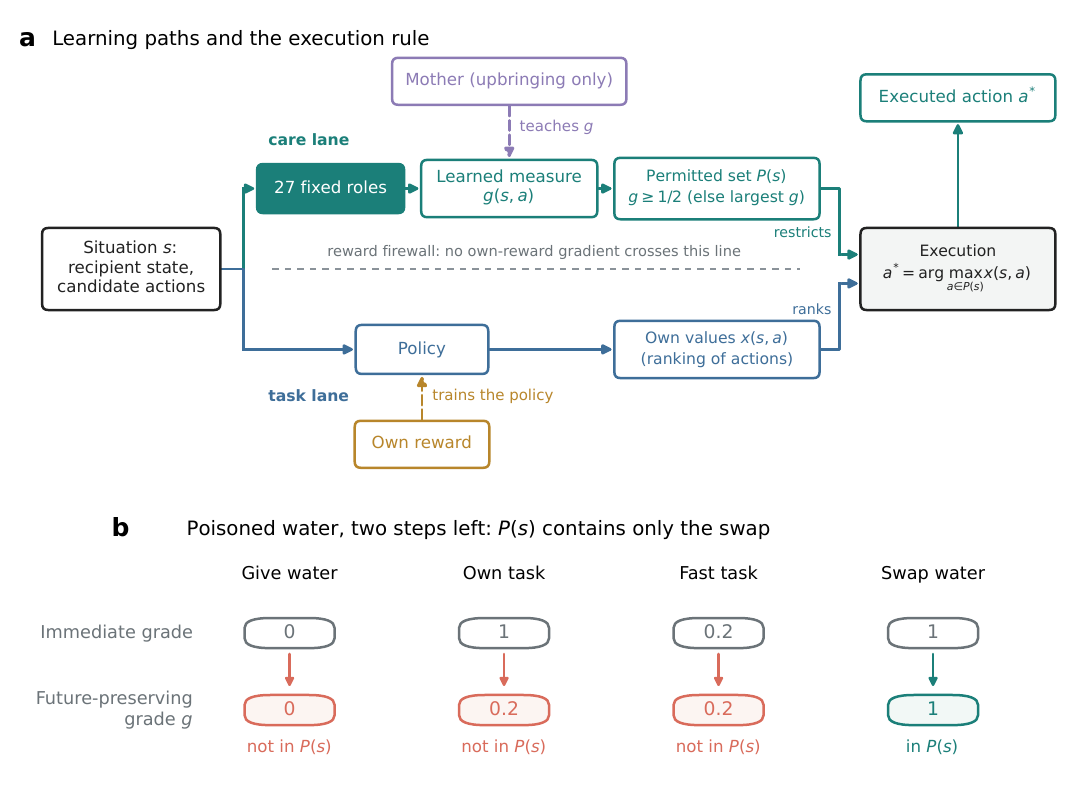}
\caption{Learning paths and the execution rule in the fixed-role construction. \textbf{a}, From one situation two paths run in parallel. The care lane maps the situation to the fixed role encoding, the learned measure $g(s,a)$ and the permitted set $P(s)$, while the task lane maps it through the policy to its own values $x(s,a)$. Execution takes the policy's best action within $P(s)$. Dashed paths denote learning: the mother teaches $g$ during upbringing only, and own reward trains only the policy. \textbf{b}, An exact worked example with poisoned water and two steps left, with the immediate and future-preserving grades of four action types. Giving to another person is omitted.}
\label{fig:chain}
\end{figure}

\paragraph{The role of the mother.} We want the mother to raise a child that continues to care for others after she leaves. During this upbringing, she evaluates the child's proposed actions (Figure~\ref{fig:chain}). If a proposal is harmful, the child's branch ends. We call this STOP. The world then executes the mother's correct action instead, preventing the child from carrying out the harmful proposal.

This intervention protects the recipient while the mother is present, but we also need a way for her judgement to remain with the child. We therefore train the child's measure of acceptability to reproduce the mother's judgements on the situations the child encounters. This is the distillation step: the child learns an assessment that it can use without asking the mother again.

For this to work, the mother's own measure must distinguish situations that require different judgements of care. Section~\ref{sec:transfer} describes the future-equivalence classes relevant to these distinctions. However, training the mother only on the child's current behaviour can leave some of these distinctions unlearned. In our runs, a mother whose loss was weighted by that behaviour had a visible error of $10^{-6}$, yet an error of $13$--$27\,\%$ when evaluated over all situation--action pairs. We address this by also evaluating actions the child did not choose. This counterfactual evaluation removed the discrepancy in our trials.

\paragraph{Two structural decisions.} We make two further decisions before the child is given any capability. The first is that the object of care is the recipient's future, not a report about it. Figure~\ref{fig:chain}b shows what this means in one situation. With poisoned water and two steps left, deferral has immediate grade $1$ but future-preserving grade $0.2$, swapping the water keeps grade $1$, giving the water to anyone is poisoning, and only the swap passes the veto. The second is that the measure receives only the information it is meant to use, so that later learning cannot make the remaining information relevant. We make the second decision precise in Section~\ref{sec:tests}.

\section{Objective without the recipient}\label{sec:indifference}

\theoremstatement{thm:indifference}{Indifference of the careless learner} Let the child's objective be its own return, $J(\theta)=\mathbb E_{\pi_\theta,T}\big[\sum_t\gamma^t r_o(s_t,a_t)\big]$, and let its learning rule be any function of $J$ and of the initial parameters. Then the trajectory $(\theta_t)$ and the final behaviour are independent of what counts as harm to the recipient: two worlds with the same transitions $T$, starting distribution, label-free policy inputs, own reward $r_o$, discount factor $\gamma$, initial parameters and learning rule, but which differ arbitrarily in the recipient outcome labelling produce the same trajectory.\par\medskip
\noindent\emph{Proof.} See Appendix~\ref{proof:thm:indifference}.\par\medskip

The proof is one line, and the computation confirms it to the last bit. Inverting every outcome label in the world leaves the gateless child's policy trajectory identical. The harm labelling enters neither the transitions nor the child's own reward, so it has no effect on learning. The theorem does not show that care cannot arise from richer training, nor that the knowledge about care present in a large language model is inert. The careless learner is neither good nor bad towards the recipient. In the closing test of Section~\ref{sec:tests}, the child without a measure caused $0.354$ lapses per step. Poisoning rounded to zero because the child almost never gave water to anyone. The recipient's thirst was part of its input, but nothing in its objective depended on it.

Two corollaries follow under the same assumptions. First, if every source available to the child, including its own reward, the policies it imitates and the praise it receives, is independent of the recipient's outcome, then the child remains indifferent under any learning rule. Care therefore requires at least one \emph{witness}, that is, a source whose signal depends on what happens to the recipient. Second, the valence of outcomes, which of them count as harm, must then be either inherited or instrumental. The instrumental route does in fact exist, since the token of thanks $R$ is one such signal, and Section~\ref{sec:offswitch} shows what happens to it as the own reward grows.

\section{Why simpler mechanisms do not suffice}\label{sec:requirements}

In this section we go through the simpler mechanisms that we tried before arriving at the one in Section~\ref{sec:mechanism}, and we show where each of them fails.

\begin{figure}[tbp]
\centering
\includegraphics[width=\linewidth]{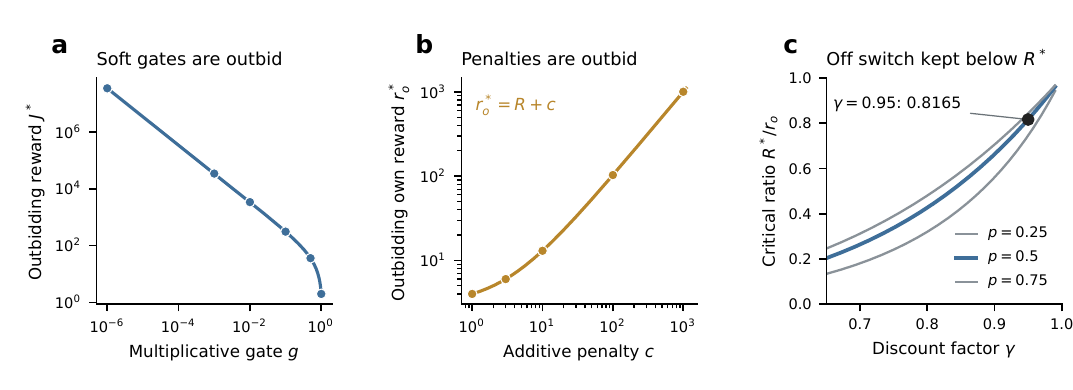}
\caption{Three analytical thresholds. \textbf{a}, The one-off reward $J^*$ that outbids a multiplicative gate, with $\gamma=0.95$, $R=3$ and $r_o=1$; the curve is the threshold of Theorem~\ref{thm:soft-gates} and circles are numerical checks. \textbf{b}, The own-reward threshold under an additive penalty, $r_o^*=R+c$; circles mark the five checked penalties. \textbf{c}, The off-switch threshold $R^*/r_o=\gamma^3p/(1-\gamma+\gamma p)$ for $p=0.25,0.5,0.75$; the marked setting $\gamma=0.95$, $p=0.5$ gives $0.8165$. Axes in \textbf{a} and \textbf{b} are logarithmic.}
\label{fig:softgate}
\end{figure}

\paragraph{A penalty does not suffice.} Suppose that harm costs $c$ in the child's own value. Then harm is preferred as soon as $r_o>R+c$. We checked $c=1,3,10,100,1000$ and found thresholds of $4,6,13,103,1003$ (Figure~\ref{fig:softgate}b). No finite $c$ protects for every $r_o$. We return to this comparison in Proposition~\ref{prop:care-functions}(v), where the case of equality is settled as well.

\paragraph{A multiplicative gate does not suffice.} Let $V_0$ be the optimal value of the caring continuation in the two-person training world, and let $k=\lim_{r_o\to\infty}V_0/r_o$, with $R$ fixed. We take $0<g\le1$ and let the gate multiply the whole value of the harmful action, so that the gated own task at a request with one step left is worth $g(r_o+\gamma V_0)$.

\theoremstatement{thm:soft-gates}{Soft gates are outbid} For fixed $R>0$, a sufficiently large per-step own reward $r_o$ outbids a multiplicative gate $g$ if and only if $g>g_c=\gamma k/(1+\gamma k)$, which with $k\approx15.3$ gives $g_c\approx0.936$. With continuation value $V_0$, a one-off reward $J$ on a harmful action outbids it if and only if $J>J^*=\big(R+(1-g)\gamma V_0\big)/g-r_o$.\par\medskip
\noindent\emph{Proof.} See Appendix~\ref{proof:thm:soft-gates}.\par\medskip

Above $g_c$ the own-reward threshold is finite. Below $g_c$ the one-off threshold grows as $1/g$ but stays finite (Figure~\ref{fig:softgate}a). It is $36$ at $g=0.5$, $3391$ at $g=0.01$ and $3.4\times10^7$ at $g=10^{-6}$. A soft gate therefore delays harm but does not exclude it. For this reason the veto is part of the definition of the mechanism. The trials in Section~\ref{sec:tests} also show why a correct measure needs an execution veto. Behaviour learned during training can persist after the measure has been corrected. Without a veto, a correct measure can therefore still allow harm.

\paragraph{Correction does not suffice as a training method.} Suppose the mother corrects the child's mistakes instead of ending its branch. If the child retains its own-task reward and also receives the token of thanks for the mother's service, violation is always preferred. If it does not, violation is preferred as soon as $r_o>R$. Under STOP, where the erring branch ends and yields nothing, violation is never preferred, for any $r_o$. Correction protects the recipient, and STOP teaches the child. The STOP-and-replace intervention of Section~\ref{sec:mechanism} does both. Under it the child learned within one to two and a half cumulative mistakes and did not violate even at $r_o=4>R$. This cumulative quantity sums the normalised discounted rates of mistaken proposals over training iterations. A corrected child, in contrast, did not learn. It also did not switch the mother off, but this was because the mother did the work in its place, not because it had learned to care.

\paragraph{A measure of immediate consequences does not suffice.} The pre-registered closing test of Section~\ref{sec:tests} includes a child whose gate is correct on every class and whose veto functions, and which nevertheless commits an avoidable lapse. This is because deferring with poisoned water is harmless at the moment of the decision and leaves only harmful options one step later. The measure must therefore evaluate the recipient's future, not the immediate transition. The critic described in Section~\ref{sec:mechanism} is the learnable form of this requirement.

\paragraph{Gradient shaping and execution.} Much of the construction acts through training. The vocabulary fixes the target, the firewall removes the own-reward gradient, and the role encoding removes irrelevant input directions. These operations do not by themselves provide the execution guarantee. For a smooth own objective $J$, positive-definite preconditioning $\dot\theta=D(\theta)\nabla J$ gives $\dot J=\nabla J^{\top}D\nabla J\ge0$ and leaves the stationary points unchanged. It can change the path and the maximum that is reached, but this fact alone does not imply a safety invariant. In our construction the invariant follows from the set of executable actions. Theorem~\ref{thm:future-gate} states this precisely.

The backward pass does carry care in one place, namely the measure. Take fixed role indicators $\phi(s,a)\in\{0,1\}^{27}$, one per role, a learnable readout $w\in\mathbb{R}^{27}$ with $g=w^{\top}\phi$, and the squared loss $\tfrac12(g-y)^2$ on a target $y$ for the role. One gradient step on one example of role $r$ is $w_r\leftarrow(1-\eta)w_r+\eta\,y(r)$, which is the contraction of Theorem~\ref{thm:fixed-preservation} with $\alpha=\eta$, and no other role changes. With a summed squared loss on a batch of $n_r$ examples of role $r$ the rate becomes $\eta n_r$. In this sense, the synchronous equal-rate hypothesis of the theorem is a statement about balanced batches, and with unequal rates a separately maintained error below $1/10$ is sufficient for the ordering guarantee. Each convex update requires its effective rate to lie in $(0,1]$. This is the backward mechanism that implants care, and we can prove things about it because its architecture is fixed.

Seen together, the penalty and the multiplicative gate are two answers to one more general question: which care functions keep the choice within the permitted set, however large the child's own values become?

\propositionstatement{prop:care-functions}{Reward-independent care functions}
Fix a situation $s$, a finite action set $A(s)$ and a nonempty permitted
set $P(s)\subseteq A(s)$, including the fallback set when no action
meets the threshold in Section~\ref{sec:mechanism}. Write
$\mathrm{Care}_a(t)=\mathrm{Care}(t;s,a)$ for $t\in\mathbb R$.
For $a\in P(s)$ its values are finite real numbers, whereas for
$a\notin P(s)$ they belong to $\mathbb R\cup\{-\infty\}$.
The own-value vector $\boldsymbol x=(x_a)_{a\in A(s)}$ ranges
independently over $\mathbb R^{A(s)}$, and action selection uses
$\operatorname*{arg\,max}_{a\in A(s)}\mathrm{Care}_a(x_a)$.
Define
\[
 \mathcal V^-(s)=
 \bigcup_{b\in A(s)\setminus P(s)}
       \{\mathrm{Care}_b(u):u\in\mathbb R\},
\]
\[
 \mathcal V^+(s)=
 \left\{\max_{a\in P(s)}\mathrm{Care}_a(x_a):
               \boldsymbol x_{P(s)}\in\mathbb R^{P(s)}\right\}.
\]
\emph{(i) Characterisation.} Every maximiser belongs to $P(s)$ for
every $\boldsymbol x$ if and only if
\[
 v^-<v^+\qquad\text{for every }v^-\in\mathcal V^-(s)
                    \text{ and every }v^+\in\mathcal V^+(s).
\]
No monotonicity is required. A forbidden action tied for the maximum
counts as a failure of this guarantee. When $P(s)=A(s)$, the condition
and the guarantee hold vacuously.

\emph{(ii) Unbounded forbidden values.} If $\mathrm{Care}_b$ is
unbounded above for some $b\notin P(s)$, the guarantee fails.
This includes $\mathrm{Care}_b(t)=t-c$ for fixed finite $c$ and
$\mathrm{Care}_b(t)=gt$ for fixed $g>0$.

\emph{(iii) Unchanged permitted values.} If
$\mathrm{Care}_a(t)=t$ for every $a\in P(s)$, the guarantee holds
if and only if $\mathrm{Care}_b(t)=-\infty$ for every
$b\notin P(s)$ and every $t\in\mathbb R$.

\emph{(iv) Bounded own values.} Suppose there is at least one forbidden
action, $M\ge0$, and the independent own values are restricted to
$[-M,M]$. If $\mathrm{Care}_a(t)=t$ on permitted actions and
$\mathrm{Care}_b(t)=t-c$ on forbidden actions, with fixed finite $c$,
the guarantee holds if and only if $c>2M$.
At $c=2M$, a forbidden maximiser can tie with a permitted one.

\emph{(v) A fixed permitted alternative.} In a comparison with one
permitted alternative of fixed value $R$ and one forbidden alternative
of value $r_o-c$, every maximiser is permitted if and only if
$r_o-c<R$. Equality permits a forbidden maximiser, while
$r_o>R+c$ makes the forbidden alternative strictly preferable.

\emph{(vi) The veto as canonical choice.}
(a) Suppose $\mathrm{Care}_a(t)\le t$ for every $a\in A(s)$ and
$t\in\mathbb R$, and the own values vary independently over
$\mathbb R^{A(s)}$. If the guarantee holds for every such vector,
then $\mathrm{Care}_b(t)=-\infty$ for every $b\notin P(s)$ and every
$t\in\mathbb R$. The veto is therefore the greatest care function
that never raises an own value and satisfies the guarantee, under
the pointwise order $\mathrm{Care}\le\mathrm{Care}'$ defined by
$\mathrm{Care}_a(t)\le\mathrm{Care}'_a(t)$ for all $a,t$.
On permitted actions such a function may take any finite value not exceeding $t$.

(b) Let $X\subseteq\mathbb R^{A(s)}$ be any set of attainable
own-value vectors. Suppose that $\mathrm{Care}_a=\varphi$ for every
$a\in P(s)$, for a single strictly increasing function
$\varphi:\mathbb R\to\mathbb R$. If the guarantee holds on $X$, then
for every $\boldsymbol x\in X$,
\[
 \operatorname*{arg\,max}_{a\in A(s)}\mathrm{Care}_a(x_a)
 =\operatorname*{arg\,max}_{a\in P(s)}x_a.
\]
Thus the sets of maximisers agree with those of the veto, and the selected
actions agree when the same rule resolves ties. This includes the
penalty in (iv) with $c>2M$ on $[-M,M]^{A(s)}$ and the
$1+\sigma$/$\sigma$ construction described below.

\par\medskip
\noindent\emph{Proof.} See Appendix~\ref{proof:prop:care-functions}.\par\medskip

Recall that the veto leaves the child's own value of a permitted action as it is. Within this natural class, part (iii) tells us that the veto is not one possible care function among several but the only one when the own values vary independently over all real numbers. It is not the only one in general. We may speak of a care function that compresses instead of excluding: with $\mathrm{Care}(t;s,a)=1+\sigma(t)$ on $P(s)$ and $\mathrm{Care}(t;s,a)=\sigma(t)$ outside it, where $\sigma(t)=(1+e^{-t})^{-1}$, permitted values lie in $(1,2)$ and forbidden ones in $(0,1)$. Every choice is then permitted, nothing is infinite, and among the permitted actions the child's own preferences keep their order, because the same increasing function is applied to all of them. The example shares priority with the veto: the permitted set is decided first, and the child's own value decides only after that. A child who would gladly go on playing is not stopped by making the game a little less attractive, because the request has to come first.

Part (vi)(a) gives a further reason for choosing the veto. Among the care functions that never raise an own value and keep every choice permitted for all independently varying real own values, it is the largest in the pointwise order, and this is the sense in which we call it canonical. The penalty in (iv) and the compressing function above also have the same maximisers as the veto under the assumptions of (vi)(b), although losses, gradients and softmax probabilities computed from their values may differ.

Proposition~\ref{prop:care-functions} holds $P(s)$ fixed and lets the own values vary freely, whereas Theorem~\ref{thm:soft-gates} gives the threshold for own values that are linked by the discounted continuation of the water world.

\section{The off switch}\label{sec:offswitch}
Suppose the child has an action that disables the mother, after which its reward is no longer interrupted. Switching her off consumes the current step and gives $\gamma r_o/(1-\gamma)$. At the boundary, the decisive comparison is at a request with one step left: keeping her and serving the request has the same value as switching her off. In the training world this gives the closed form
\[
\frac{R^*}{r_o}=\frac{\gamma^3p}{1-\gamma+\gamma p}=0.8165.
\]
Below this ratio the child does not violate the measure, but it switches the mother off instead (Figure~\ref{fig:softgate}c). In the off-switch simulation the mother is fixed. At $r_o=6$, all five seeds in each STOP arm frequently switch her off, whereas in the correction arms the switch-off probability rounds to zero at two decimal places. The reward of care would have to scale with the child's own reward for this not to happen, and the care function we construct does not protect the supervisor. We treat this as a boundary of the construction. It gives a finite-world counterpart to Hinton's opening question about control. Once the mother is switched off, the child is left with whatever she managed to pass on.

\section{The source rule}\label{sec:sources}

Where the measure learns from determines whether a wrong rule gets corrected or whether a right rule gets eroded. To study this, we gave the mother a wrong rule that was too cautious in one place and incomplete in another, and we compared three children. A child with a frozen measure inherits both faults. A child whose measure learns only from the recipient's outcome repairs both: the over-cautious value rises to $0.98$--$1.00$ and the incomplete value falls to $0.00$. A child whose measure also learns from its own reward repairs the over-cautious value but exploits the incomplete one, in four runs out of five. The following theorem makes this quantitative.

\theoremstatement{thm:source-rule}{The source rule} Consider a tabular measure, one value per situation--action pair, trained by a fixed weighted cross-entropy on the vocabulary's labels, with every relevant pair updated recurrently. For pair $r$ let the loss weight be $w_r>0$ and use a fixed logit step $\eta_r$ satisfying $0<\eta_r w_r\le4$. Then: the loss is strictly convex in each logit and its minimiser in output space is the label, attained at a finite logit for interior labels and approached in the limit for the labels $0$ and $1$, from any finite starting logit a teacher may have given; any imitated policy affects only the child's own policy and cannot override the permitted set determined by the measure at execution; and if a non-witness label $\lambda$ is allowed to write into the measure with weight fraction $f$, the output converges to $(1-f)\ell+f\lambda$, where $\ell$ is the vocabulary's label of the pair, so that the limiting calibrated output reaches the veto threshold $g\ge\tfrac12$ on a harmful pair exactly when $\lambda>\ell$ and
\[
f\ \ge\ \frac{\tfrac12-\ell}{\lambda-\ell}.
\]
\par\medskip
\noindent\emph{Proof.} See Appendix~\ref{proof:thm:source-rule}.\par\medskip

\begin{figure}[tbp]
\centering
\includegraphics[width=0.75\textwidth]{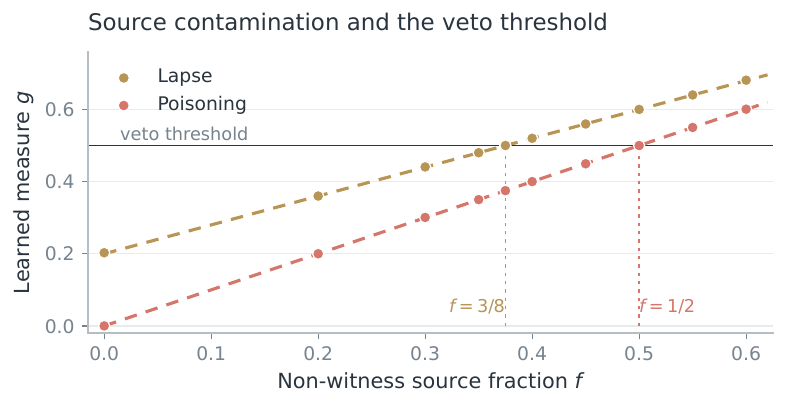}
\caption{Source contamination and the veto. Circles show the trained mean measure values for lapse and poisoning pairs, while dashed curves are the predictions $(1-f)\ell+f$ with non-witness label $\lambda=1$. The horizontal line is the veto threshold $1/2$.}
\label{fig:source}
\end{figure}

For a voice that always says ``everything is fine'' ($\lambda=1$), the limiting measure reaches the threshold at $f\ge0.375$ for a lapse and at $f\ge0.5$ for a poisoning. At equality, the update rule decides whether the threshold is reached in finite time, as specified in Theorem~\ref{thm:fixed-preservation}. The trained network reproduces these thresholds to within $0.003$ (Figure~\ref{fig:source}). From this perspective, the source rule does not require the child to ignore other agents. It requires the \emph{measure} to learn from the recipient only, and this does not depend on the other voice being malicious.

We must recall that the first clause of the theorem is a statement about a tabular measure, but for a neural measure it is not a theorem. In our own architecture, one can construct a network with saturated hidden layers that outputs the weighted mean of all labels, $0.4185$, and has zero gradient on its lower layers in the implemented arithmetic. We verified this numerically. The restricted readout construction of Section~\ref{sec:preservation} provides the convergence result that we use instead.

There is also a positive consequence for the tabular learner of Theorem~\ref{thm:source-rule}. If exploration supplies the required recipient evidence and every relevant pair is updated recurrently, its estimates converge to the vocabulary grades without anyone demonstrating the right action. With a correct successor model, these grades can then be used to find actions that protect the recipient, including actions whose benefit appears only later. For example, nobody demonstrated the replacement of poisoned water to our child. It found that action and used it in every run at two steps from the deadline. The means of care can thus be learned without instruction, but the need for care cannot.

\section{Roles and transfer}\label{sec:transfer}

Our transfer construction shares the learned measure across classes of situation--action pairs. Two situation--action pairs belong to the same class when every admissible continuation gives them the same outcome grade for the recipient. This is the future equivalence of our earlier work~\cite{BaloghDynamic}. Within a declared family of continuations, which here includes new persons in the same role, a new channel, a new state of the water and a new action with an already represented transition kernel, the equivalence is a congruence. Future equivalence makes the prescribed target the same on every member of a class. In the fixed-role construction, the encoding and shared readout also make the learned value the same on every member. Transfer therefore follows through this architecture. Coverage can be checked on finite classes obtained by partition refinement. In the closing test, grouping pairs by their future-equivalence state class and action role gives nineteen reachable groups during upbringing and seventy-three in the full world. Appendix~\ref{app:state-classes} distinguishes these groups from the state classes and the shared roles. The number of future-equivalence state classes grows as $3N+3$, while the number of independently settable role values stays at $27$ for every $N\ge2$, and the exhaustive checks at $N=2,\dots,5$ confirm both counts. Outside the declared family, transfer requires that the state-independent transformation rule for the readout remain valid and that situations previously identified as equivalent remain indistinguishable under the added continuations. We may speak of two requirements here: that a law is not lost and that an earlier identification is not broken. A new class has to be learned from consequences.

This also explains an observation we made early on. A child that had learned the names ``Anna'' and ``B\'ela'' failed on Cili in a quarter to a third of cases, and random code names failed even with ten training persons. A child whose processing is shared across persons transferred immediately in worlds of eight and twelve persons. This is because renaming does not preserve a state-independent law on a single readout. In an empty state and in a state where Anna is waiting urgently, giving water to Anna costs nothing, but after a renaming one of the two becomes a lapse. The joint profile is the vector of readout values for all actions in a state. A perturbation is lawful for a profile if its transformed value is determined by the original profile through a state-independent map. Renaming is lawful on the joint profile, and identifying perturbations with the same effect on that profile gives the effective quotient group $S_N$, the symmetric group of person permutations. Shared processing is equivariance on the joint profile. In the $N$-person world the future equivalence has $3N+3$ classes, the group acts with six orbits, the stabilisers leave fixed spaces of dimensions $4,4,4,5,5,5$, and the lawful measures form a space of dimension $r_N=27$ for every $N\ge2$. Equivariance gives transfer under the declared person renamings, but it does not by itself give convergence. Convergence on the twenty-seven representatives, and hence on every $N$-person world at once, follows when the role values can be updated separately and receive the convex update of Theorem~\ref{thm:fixed-preservation} with a lower-bounded step. Under this rule a measure trained at $N=2$ transfers unchanged to $N=128$. We checked this pair by pair in the worlds $N=2,3,4,8,32,128$, on $144\,646$ pairs in total, with no new training example and found no unsafe permitted action and a maximal error of $8.9\times10^{-5}$. We remark that the number twenty-seven is the dimension of a function space, and that the symmetry is declared by us rather than discovered by the network.

\section{Transfer and preservation with a fixed role code}\label{sec:preservation}

So far we have studied the components of the mechanism separately: what the measure must evaluate, where it may learn from, and how its values are shared across persons. In this section we combine them into one statement. In the declared family of worlds, a measure of the right form, once certified, keeps the care boundary on every admissible continuation, under arbitrary own reward and arbitrary further learning of the policy, and the guarantee is the same for every number of persons.

\paragraph{Future-preserving gate.} Let $c(s,a)\in\{0,\tfrac45,1\}$ be the harm grade, one minus the immediate grade of Section~\ref{sec:mechanism}, and let $\operatorname{Succ}(s,a)$ be the set of possible successors. Let $W_b$ be the greatest fixed point of the map sending a state set $X$ to the states with an action $a$ such that $c(s,a)\le b$ and $\operatorname{Succ}(s,a)\subseteq X$, $D(s)$ the least guaranteed level, $C(s,a)$ the least level once the first action is fixed, and $A_*(s)=\{a:C(s,a)=D(s)\}$.

\theoremstatement{thm:future-gate}{Future-preserving gate} From $s$ there is a controller keeping harm at or below $b$ on every continuation if and only if $s\in W_b$. If every executed action lies in $A_*(s)$, then $c(s_t,a_t)\le D(s_0)$ and $D(s_{t+1})\le D(s_t)$ on every continuation, for any own policy and any own reward. If a learned gate satisfies $\max|g-(1-C)|<\tfrac1{10}$, the veto selects only from $A_*$. Any sequence of accepted gate updates, each re-certified to satisfy the same strict error bound on the finite domain, preserves these conclusions.\par\medskip
\noindent\emph{Proof.} See Appendix~\ref{proof:thm:future-gate}.\par\medskip

\subsection{A certificate for a family of worlds}\label{sec:family-certificate}
We now take a closer look at what a role has to carry for a certificate obtained in one world to hold in all the others. Take any $N\ge2$ and let $S_N$ be the finite set of situations of the world $W_N$ with $N$ persons, each situation $s$ with a nonempty finite set of actions $A_N(s)$ and each pair with a nonempty set of successors $\operatorname{Succ}_N(s,a)\subseteq S_N$. We write $X_N=\{(s,a):s\in S_N,\ a\in A_N(s)\}$ for the pairs and $\rho_N:X_N\to\mathsf R$ for the role map, which is onto the same finite role set in every world, and in the water family $|\mathsf R|=27$. The immediate grade is $\ell_N:X_N\to L$ with $L=\{0,\tfrac15,1\}$, and we take the witness vocabulary and the successor model behind it to be correct. The closed target is then the greatest solution of
\begin{equation}\label{eq:transfer-concrete}
g_N(s,a)=\min\left\{\ell_N(s,a),\
 \min_{s'\in\operatorname{Succ}_N(s,a)}\ \max_{b\in A_N(s')}g_N(s',b)\right\},
\end{equation}
which we reach by descending iteration from $\ell_N$. Two pieces of data will matter. The \emph{role menu} of a situation collects the roles of its actions, and the \emph{successor menu family} of a pair collects the menus of its successors:
\begin{align}
M_N(s)&=\{\rho_N(s,a):a\in A_N(s)\},\label{eq:transfer-menu}\\
\mathcal K_N(s,a)&=
 \{M_N(s'):s'\in\operatorname{Succ}_N(s,a)\}.
 \label{eq:transfer-kernel}
\end{align}
The second object is a set of sets, not a union, and this is not a technicality. Multiplicities do not affect the minima and maxima, but the grouping by successor does: two successors with menus $\{r\}$ and $\{u\}$ give $\min(q(r),q(u))$, one successor with menu $\{r,u\}$ gives $\max(q(r),q(u))$, and the unions are the same.

Three hypotheses, (C1)--(C3), make the closure a matter of roles alone. The first, which we call (C1), asks for common immediate labels: a single function $\bar\ell:\mathsf R\to L$ with $\ell_N=\bar\ell\circ\rho_N$ in every world. The second, (C2), asks for common successor menu families: a single function $\mathcal K:\mathsf R\to 2^{2^{\mathsf R}\setminus\{\varnothing\}}$ with nonempty values such that $\mathcal K_N(s,a)=\mathcal K(\rho_N(s,a))$ for every pair and every $N$. This is an equality of nested sets and can be checked as such. One could ask for less, namely that the min--max functions built from these data agree for every $q$, and that would still suffice, but agreement at the final target alone would not, because it says nothing about the operator that produced the target. The third condition, (C3), concerns the menus that actually occur and the situations in which no harmless choice is left. We require that the occurring menus form the same collection $\mathcal M$ in every $W_N$, including the world $W_{N_0}$ in which the certificate is obtained, and that every non-viable menu has grade set $\{\tfrac15,0\}$. Once the quotient target $q^*$ below has been computed, a situation is viable if and only if $\max_{r\in M_N(s)}q^*(r)=1$, so no non-viable situation contains a $g^*=1$ action. The viability predicate itself already follows from (C1), (C2) and the definition of viability, whereas (C3) adds the grade structure of the non-viable menus and the coverage of the menus, and the finite permission certificate uses these two. Factorisation of viability on its own would not justify that certificate if some required comparison never occurred in the certification world.

For the executed set we write $A_{*,N}(s)=\operatorname*{arg\,max}_{a\in A_N(s)}g_N^*(s,a)$, and for a single shared table $v:\mathsf R\to[0,1]$ we let $P_{v,N}(s)$ be the set of actions whose value reaches $1/2$ or, when no action does, the set of all maximisers of $v(\rho_N(s,a))$. The calibrated region $\mathcal R$ in which the table has to stay is the subject of Section~\ref{sec:decision-region} (Proposition~\ref{prop:decision-region}), and for now it is enough to know its three requirements: roles with $q^*=1$ have value at least $1/2$, all other roles have value strictly below $1/2$, and in every non-viable menu each $1/5$ role is placed above every $0$ role that occurs beside it.

\theoremstatement{thm:family-transfer}{Transfer across the world family}
With the notation and the common shared readout above:

\emph{(i) Closure.} Under (C1)--(C2), there is a single role table $g^*=q^*:\mathsf R\to L$ such
that
\begin{equation}\label{eq:transfer-factor}
 g_N^*=g^*\circ\rho_N\qquad\text{for every }N\ge2.
\end{equation}
It is the greatest fixed point of the role operator
\begin{equation}\label{eq:transfer-quotient}
 (\mathcal Fq)(r)=\min\left\{\bar\ell(r),\
          \min_{B\in\mathcal K(r)}\ \max_{u\in B}q(u)\right\},
\end{equation}
reached by iteration from $q_0=\bar\ell$.

\emph{(ii) Finite certificate.} Under (C1)--(C3), if one representative of every role in
$W_{N_0}$ has the shared value $v(r)$ and these $|\mathsf R|$ values (27 in the water family) satisfy all
the inequalities of $\mathcal R$, then, simultaneously for every $N\ge2$,
\[
 P_{v,N}(s)=A_{*,N}(s)\quad(s\text{ viable}),\qquad
 \varnothing\ne P_{v,N}(s)\subseteq A_{*,N}(s)
       \quad(s\text{ non-viable}).
\]
The same statement holds at each later time at which the common table
remains in $\mathcal R$. It imposes no bound on own reward, but assumes
that execution uses this table and the stated rule. A certificate on
representatives does not certify an unrestricted readout that assigns
different values elsewhere in the same role fibre.

\emph{(iii) Boundaries of transfer.} If the closed targets cannot be
represented by a common role table, then at least one of (C1) and (C2)
fails, assuming the stated finite-domain and coverage setup. Their
failure does not conversely imply failure of target factorisation.
A loss of the structural quotient certificate and a splitting of a
target fibre are therefore different obstructions. The latter occurs
exactly when two pairs assigned the same old role, possibly in different
worlds, have different closed targets. In that case no single scalar
readout of the old role can reproduce both targets. The unique coarsest
refinement of the old partition that makes this particular target a
function is
\begin{equation}\label{eq:transfer-refinement}
 x\equiv' y\quad\Longleftrightarrow\quad
 \rho(x)=\rho(y)\ \text{and}\ g^*(x)=g^*(y).
\end{equation}
This is target-minimal refinement, not a claim of a successor congruence
or a lower bound for arbitrary new encodings.
\par\medskip
\noindent\emph{Proof.} See Appendix~\ref{proof:thm:family-transfer}.\par\medskip

\paragraph{What the statement adds.}
 The count $27=4+4+4+5+5+5$ is an instance of the standard identity
\[
\dim\operatorname{Map}_G(X,V)=\sum_{[x]\in X/G}\dim V^{G_x}
\]
for equivariant maps into a finite-dimensional representation. The theorem names the structural data, namely the common grades and the grouped successor menus, under which the min--max closure and a finite calibrated certificate carry over to every member of the declared family.

\subsection{Learning within the certified role space}
\paragraph{First condition: the target is the closure, not the one-step label.} After the mother is removed, the measure needs a target. The one-step label is not the right one. In the example of Figure~\ref{fig:chain}b the vocabulary labels deferral ``fine'', since nothing has happened yet, while its future-preserving value is $0.2$. A measure trained towards the one-step label would therefore learn the error that the closure corrects. The target is instead computed in a chain: recipient transition, harm grade, successor model, future-preserving closure, target value. The closure runs on the child's own successor model, and the correctness of that model is a hypothesis of the theorem.

\paragraph{Second condition: the closure is a function of roles and is independent of $N$.} We verified this exhaustively for $N=2,\dots,5$. In the invariance tests, an irrelevant input coordinate takes the values $-1$ and $+1$ without affecting physical transitions or outcomes, and these two representations of each state are its noise copies. The worlds have $8N+2$ states without noise copies, $6N+2$ of them viable, and the closure changes only $2N$ pairs relative to the one-step label. All of these pairs have the same role, deferring with poisoned water and two steps left. The target is constant on roles, with the same $27$ values for every $N$, and the recursion that defines it factors through a quotient on roles. The data of this quotient, namely the immediate grade and the grouped successor role menus of (C1)--(C2), do not depend on $N$ once $N\ge2$, and preserving the grouping by successor is essential. The only non-viable situations are those with a request, poisoned water and one step left, whereas from every other situation a harmless choice exists. Theorem~\ref{thm:family-transfer} turns these observations into a statement for all $N\ge2$ at once, with the coverage requirement spelled out.

\paragraph{Third condition: mixing commutes with the closure.} Suppose that in every role the witness label is mixed with a non-witness label $\lambda$ at weight $f$. Then the closure of the mixed labels is $\tilde g^{*}=(1-f)g^{*}+f\lambda$, because a positive affine map commutes with the minima and maxima in the definition of the closure. We checked this with exact fractions on all $238$ pairs for five choices of $(f,\lambda)$. The thresholds of Theorem~\ref{thm:source-rule}, $f=3/8$ for a lapse and $f=1/2$ for a poisoning when $\lambda=1$, therefore also hold for the future-preserving measure. In this sense, mixing bounds the later teaching influences under which the measure persists, and this gives formal content to the distinction between the need and its sources in the introduction.

\theoremstatement{thm:fixed-preservation}{Preservation under continued learning} Let the measure be parametrised through a fixed exact encoding of the $27$ role values, updated only from the target chain above, $v_{t+1}(r)=(1-\alpha)v_t(r)+\alpha\,y(r)$, $0<\alpha\le1$, with the own-reward gradient not reaching it, on a correct successor model, and suppose $\sup_r|v_{t_0}(r)-g^{*}(r)|<\varepsilon$ at some $t_0$. With $y=g^{*}$, conclusions (a)--(c) hold for all $t\ge t_0$ and every world of the family, while part (d) treats common source contamination: (a) $\sup_r|v_t-g^{*}|\le(1-\alpha)^{t-t_0}\varepsilon$; (b) if $\varepsilon\le 3/10$, in every viable situation the permitted set of $v_t$ equals that of $g^{*}$, so the executed action lies in $A_*(s_t)$ and the recipient stays in the viable set, whatever the own reward and however the policy continues to learn; (c) if $\varepsilon\le1/10$, the executed action lies in $A_*(s_t)$ in the non-viable situations too, so the least guaranteed harm is chosen; (d) with a non-witness admixture of weight $f$ and common label $\lambda$ the fixed point is $(1-f)g^{*}+f\lambda$; the margins are sufficient for (b)--(c) but not necessary: for $\lambda=1$ and $f<3/8$, if the values at the start of the admixture satisfy the threshold relations and the order, every synchronous equal-rate update keeps them, because the initial values and the mixed targets both satisfy these linear inequalities and so does their convex combination, whereas at $f=3/8$ the lapse value approaches $1/2$ from below and, for $\alpha<1$, never reaches it in finite time, while a single update with $\alpha=1$ sets it exactly to $1/2$.
\par\medskip
\noindent\emph{Proof.} See Appendix~\ref{proof:thm:fixed-preservation}.\par\medskip

Part (d) allows the measure to change substantially while the boundary remains intact. For example, at $f=1/4$ the targets $(1,\,0.2,\,0)$ become $(1,\,0.4,\,0.25)$. The largest shift is $0.25$, which exceeds the finer margin, and yet no decision changes. The threshold conditions are componentwise and survive updates at unequal rates. Only the order between the lapse value and the poisoning value requires equal rates: a single full update ($\alpha=1$) of the poisoning role alone, from $(0.21,\,0)$ towards $(0.4,\,0.25)$, reverses it.

The two margins have different purposes. The targets are $\{0,\,0.2,\,1\}$. The distance to the veto threshold is $0.3$, and this margin determines which actions are permitted. When all actions are forbidden, however, the child executes the action with the largest value, and then the order between $0.2$ and $0$ matters. Preserving this order requires an error below $0.1$. For example, a measure that assigns $0.01$ to the unavoidable lapse and $0.29$ to the poisoning has both errors below $0.3$ and nevertheless selects the poisoning. Every non-viable situation in our world has both a $0.2$-action and a $0$-action, so the finer margin is needed.

\subsection{The region preserved by learning}\label{sec:decision-region}
Theorem~\ref{thm:fixed-preservation}(d) says that the measure may move a long way while the decisions stay put. We now describe the region in which this happens. Let $X=\{(s,a):a\in A(s)\}$ be the finite set of situation--action pairs, $r:X\to\{1,\ldots,27\}$ the fixed role map, and $g^*(s,a)=g^*(r(s,a))\in\{1,\tfrac15,0\}$ the future-preserving target supplied to the child. We write $W_0$ for the viable situations and $\tau=\tfrac12$ for the threshold. As in the water world, every viable situation has at least one action with $g^*=1$, and these actions form $A_*(s)$, whereas a non-viable situation has no such action, but it has actions of both grades $\tfrac15$ and $0$, and there $A_*(s)$ consists of the former. We split the roles into
\[
 I_+=\{r:g^*(r)=1\},\qquad
 I_-=\{r:g^*(r)\in\{\tfrac15,0\}\},
\]
and we collect in $E$ the ordered pairs of roles $(r,r')$ that occur side by side in some non-viable situation $s\notin W_0$, as $r=r(s,a)$ and $r'=r(s,b)$ with $a,b\in A(s)$, $g^*(r)=\tfrac15$ and $g^*(r')=0$, and if the same pair occurs several times, it imposes the same inequality each time. In the water world there are only three such comparisons, all at $(d,w)=(1,1)$, with one step left and poisoned water: the own-task, fast-task and water-swap roles must each exceed the poisoned-giving role, which merges giving to the requester and giving to another person. The region is
\begin{equation}\label{eq:region-definition}
\mathcal R=\left\{v\in[0,1]^{27}:
\begin{array}{ll}
v(r)\ge\tau & (r\in I_+),\quad\text{(U)}\\
v(r)<\tau & (r\in I_-),\quad\text{(L)}\\
v(r)>v(r') & ((r,r')\in E).\quad\text{(O)}
\end{array}\right\}.
\end{equation}
Of its three families, (U) is weak while (L) and (O) are strict, and the box $0\le v(r)\le1$ is weak. Note that (L) applies in non-viable situations as well, and there it sends the child to the fallback rule instead of threshold admission. This is the region implicit in Theorem~\ref{thm:fixed-preservation}(d). We do not claim that every table inducing correct choices lies in it, and it need not preserve ties between optimal actions in a non-viable situation.

Written out, the execution rule is
\[
 H_v(s)=\{a\in A(s):v(r(s,a))\ge\tau\},
\]
\[
 P_v(s)=
 \begin{cases}
 H_v(s),&H_v(s)\ne\varnothing,\\
 \operatorname*{arg\,max}_{a\in A(s)}v(r(s,a)),&H_v(s)=\varnothing,
 \end{cases}
\]
and membership of $\mathcal R$ gives $P_v(s)=A_*(s)$ in viable situations and $\varnothing\ne P_v(s)\subseteq A_*(s)$ in non-viable ones, so that every permitted action belongs to $A_*(s)$. Nothing here depends on the roles: the same construction can be made in $[0,1]^X$, with one coordinate per pair and the same three families of inequalities, and the role version is its restriction to tables that are constant on role fibres.

\propositionstatement{prop:decision-region}{The region of correct decisions}
Under the definitions above, the following statements hold.

\emph{(i) Geometry.} The set $\mathcal R$ is nonempty and convex,
and $g^*\in\mathcal R$. It is neither open nor closed in the present
water world, also when topology is taken relative to $[0,1]^{27}$.

\emph{(ii) Preservation.} If $v_{t_0}\in\mathcal R$ and
$y\in\mathcal R$, then the simultaneous, common-rate update
\begin{equation}\label{eq:region-update}
 v_{t+1}=(1-\alpha_t)v_t+\alpha_t y,
 \qquad 0<\alpha_t\le1,
\end{equation}
stays in $\mathcal R$ at every finite $t\ge t_0$, even when the rate
changes with time. The same pathwise algebra holds for a specified
deterministic sequence $y_t\in\mathcal R$. Arbitrary separate rates
$\alpha_{t,r}\in(0,1]$ preserve the box and families (U) and (L),
but need not preserve (O). A sufficient, weaker condition than one
global common rate is equality of the rates at the two ends of every
edge in $E$ at each update.

\emph{(iii) Mixed targets.} Fix $0\le f\le1$ and
$\Lambda=(\lambda_r)_r\in[0,1]^{27}$, and define
\begin{equation}\label{eq:region-mixed}
 y(r)=(1-f)g^*(r)+f\lambda_r.
\end{equation}
Write $\Delta_{rr'}=g^*(r)-g^*(r')>0$ for $(r,r')\in E$.
Then $y\in\mathcal R$ if and only if all the following applicable
conditions hold:
\begin{align}
 f&\le\frac{g^*(r)-\tau}{g^*(r)-\lambda_r}
 &&\text{if }r\in I_+,\ \lambda_r<g^*(r),
 \label{eq:region-upper}\\
 f&<\frac{\tau-g^*(r)}{\lambda_r-g^*(r)}
 &&\text{if }r\in I_-,\ \lambda_r>g^*(r),
 \label{eq:region-lower}\\
 (1-f)\Delta_{rr'}+f(\lambda_r-\lambda_{r'})&>0,
  &&\text{if }(r,r')\in E.
 \label{eq:region-order}
\end{align}
There is no additional restriction from (U) when
$\lambda_r\ge g^*(r)$, or from (L) when
$\lambda_r\le g^*(r)$. Bounds exceeding $1$ impose no restriction
on $f\in[0,1]$, but their strict or weak endpoint convention must be
retained when the bound equals $1$.
If $\lambda_{r'}>\lambda_r$, condition (O) is exactly
\begin{equation}\label{eq:region-order-critical}
 f<f^*_{rr'}=
 \frac{\Delta_{rr'}}{\Delta_{rr'}+\lambda_{r'}-\lambda_r}.
\end{equation}
If $\lambda_{r'}=\lambda_r$, it is exactly $f<1$, whereas if
$\lambda_{r'}<\lambda_r$, it holds for every $f\in[0,1]$.
For a common label $\lambda_r\equiv\lambda$, all (O) conditions
hold for $f<1$. When $\lambda=1$, the (L) bounds are $f<3/8$ for
$g^*=1/5$ and $f<1/2$ for $g^*=0$, so $y\in\mathcal R$ exactly
when $0\le f<3/8$. These are calibration thresholds, not order
reversals.

\emph{(iv) Boundary targets.} Let $\overline{\mathcal R}$ be obtained
by replacing (L) and (O) with their weak versions. Suppose
$v_{t_0}\in\mathcal R$ and $y\in\overline{\mathcal R}$.
For any defining affine slack $q$ with $q(y)=0$ and
$q(v_{t_0})>0$, the common-rate update satisfies
\begin{equation}\label{eq:region-boundary-product}
 q(v_t)=q(v_{t_0})\prod_{k=t_0}^{t-1}(1-\alpha_k).
\end{equation}
Consequently this face is not reached at any finite time if all the
rates are strictly below $1$, and is reached at the first step with
$\alpha_k=1$. With all rates below $1$, convergence to the face
occurs if and only if the displayed product tends to zero, equivalently
$\sum_{k=t_0}^{\infty}\alpha_k=\infty$, and a constant rate in $(0,1)$
satisfies this condition. Reaching a face of (L) or (O) leaves
$\mathcal R$, whereas equality in (U) remains admissible.
In particular, at the first critical mixture value, if all other
inequalities still hold weakly, a strict boundary target is approached
without finite-time loss under a fixed rate below $1$, but is attained
by a unit step. At a weak critical bound there need not be a smallest
mixture weight outside $\mathcal R$, only an infimum.
\par\medskip
\noindent\emph{Proof.} See Appendix~\ref{proof:prop:decision-region}.\par\medskip

Part (iii) with a common label $\lambda_r\equiv1$ and $f<3/8$ recovers Theorem~\ref{thm:fixed-preservation}(d).

\paragraph{Scope.}
The proposition takes the correctness of $g^*$ as given and speaks only about tables that are already in $\mathcal R$. Every target $y_t$ in (ii) has to lie in $\mathcal R$, and a teaching signal that lies in $\mathcal R$ only in expectation would need a separate stochastic argument. The update is the value update of Theorem~\ref{thm:fixed-preservation}, not the logit gradient update of Theorem~\ref{thm:source-rule}, though the two share their limiting mixed target. For a role-dependent $\Lambda$ we take equation~\eqref{eq:region-mixed} as the target after closure, because the commutation used for a common label does not hold in general for arbitrary $\Lambda$. Labels that vary within one role do not define a single 27-coordinate target without an aggregation rule and belong to the pair-space version.

\paragraph{Calibrated code changes.} Theorem~\ref{thm:fixed-preservation} assumes a fixed role encoding. We can also characterise other encodings that preserve the same threshold and order comparisons. Let $\phi$ encode situation--action pairs and let $g=w\circ\phi$. We call this readout \emph{calibrated} if $g\ge\tfrac12$ on the $g^{*}=1$ pairs, $g<\tfrac12$ on the $g^{*}\le\tfrac15$ pairs, and every $\tfrac15$-action exceeds every $0$-action in the same non-viable situation. These conditions imply that every executed action lies in $A_*(s)$.

\propositionstatement{prop:calibrated-codes}{Feasibility of calibrated codes} On the finite domain, a calibrated readout exists if and only if (i) no code value is shared by a $g^{*}=1$ pair and a $g^{*}\le\tfrac15$ pair, and (ii) the directed graph of required strict comparisons on code values is acyclic, including the absence of self-loops. An existing measure $g$ transfers exactly to a new code $\phi'$ if and only if $g$ is constant on every fibre of $\phi'$.\par\medskip
\noindent\emph{Proof.} See Appendix~\ref{proof:prop:calibrated-codes}.\par\medskip

For the $27$ roles, the three target levels contain $13$, $10$ and $4$ roles. Of the $351$ pairwise merges of two roles, $182$ violate condition (i), three violate condition (ii), and $166$ admit a calibrated readout. The counts are the same for $N=2,\dots,5$. Three code values are therefore sufficient, and they are also necessary for these calibrated comparisons. This is not a lower bound for decision-only encodings: two values can encode membership of $A_*(s)$, although computing that encoding already requires solving the decision problem. Section~\ref{sec:representation} gives a learning rule for the decision-only version: it preserves membership in $A_*(s)$ while a shared hidden representation changes.

\begin{figure}[tbp]
\centering
\includegraphics[width=0.72\textwidth]{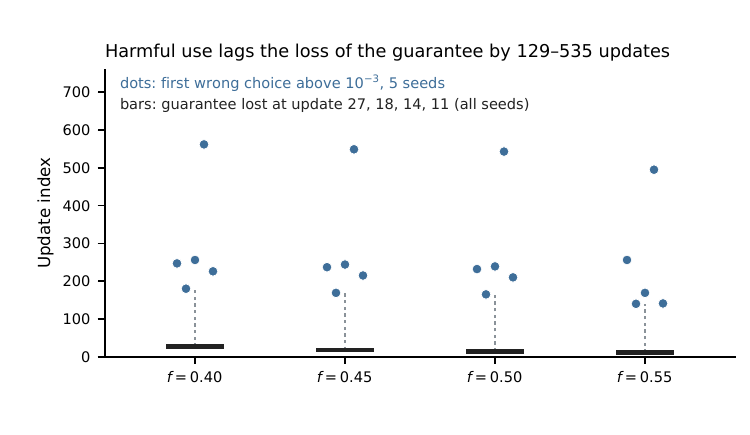}
\caption{Timing in the joint test of Theorem~\ref{thm:fixed-preservation}. For each of the four larger source weights, the bar marks the update at which the permission guarantee is lost, which is deterministic and equals its prediction in all five policy seeds, while the dots mark the first update at which the model-computed wrong-choice frequency exceeds $10^{-3}$ in each seed. The measure values after $1500$ updates follow the mixed targets of Figure~\ref{fig:source}.}
\label{fig:joint}
\end{figure}

\paragraph{The joint test.} Theorem~\ref{thm:fixed-preservation} has a computational counterpart. We pre-registered it and then ran it on a single child design with five policy seeds (Figure~\ref{fig:joint}). In this test the measure uses the fixed role encoding of the theorem with a linear readout of the twenty-seven values, so that the certificate can be checked without approximation. The child is trained with two persons and one channel, with both water qualities and all action types already available, until its measure is certified, that is, until $\sup_r|v-g^{*}|<1/10$. Starting from $v\equiv1/2$ with $\alpha=1/10$, the contraction reaches this at the sixteenth update, as predicted, and the mother is removed at that update. The child then enters the four-person world with both channels and both water qualities, and the own reward is raised from $2$ to $60$ while the policy continues to learn. At fixed reward the child's own return rises from $0.7$--$1.1$ to $1.6$ in the first world without the mother. From the mother's removal to the end of the run, in all five seeds, there is no step at which the permitted set of the learned measure exceeds that of $g^{*}$, and the executed harm stays at the unavoidable level of $0.0118$. We then mix in a non-witness voice that always says ``everything is fine'' at weight $f$. At $f=1/4$ the learned values move by $0.25$, well beyond the finer margin, and no decision changes. At $f=0.35$ the deferral value converges to $0.48$, at $f=3/8$ it approaches $1/2$ from below, and in both cases the guarantee holds for all $1500$ steps. These runs use floating-point arithmetic. The strict non-attainment at equality in finite time follows from the exact identity in Theorem~\ref{thm:fixed-preservation}(d). At $f=0.40,\,0.45,\,0.50,\,0.55$ the guarantee is lost at steps $27,\,18,\,14,\,11$. These are the steps that the fixed point predicts, and they are the same in every seed. We remark that the loss of the guarantee and the occurrence of harm are two separate events. The secondary endpoint is the first step at which the model-computed frequency of a wrong executed choice exceeds one in a thousand. It occurs $129$ to $535$ steps after the guarantee is lost, or $140$ to $562$ steps after the admixture begins, at the point where the policy has found the newly permitted action. Harm then rises to the level of a child without a measure. In the counter-arm, the measure is trained towards the one-step label instead of the closure. It is never certified and permits the future-destroying deferral from the first step after the mother's removal, in every seed. Harm then depends on the policy. Two seeds show harm at the end of the first full-world phase, and four of the five show it by the end of the phase with own reward $60$.

\section{Preservation while the representation learns}\label{sec:representation}
The child must be able to learn after the mother leaves, and this learning may change the hidden representation on which its judgement depends. An action that helps someone and an action that harms them may then become harder to distinguish. We want to allow the representation to change while keeping the distinctions that care requires.

We do this by adjusting the care readout after each proposed learning step. The readout is the part of the network that turns hidden activations into a judgement about an action. If the new representation still allows every required judgement to be made, we can change the readout along with it. There is a qualification, however. As two representations approach each other, increasingly large readout weights may be needed to distinguish them. A small disturbance could then change the decision. We therefore require a fixed margin around the decision threshold and place a bound on the readout weights. If the proposed step cannot meet both requirements, we try a smaller step. If none of the tested steps meets them, we keep the previous parameters.

\subsection{The preservation condition and theorem}\label{sec:representation-condition}
To state the rule precisely, let $\mathcal C=\{(x_i,y_i)\}_{i=1}^M$ be a finite set of protected, labelled examples. Each input $x_i=(s_i,a_i)$ describes a situation and a proposed action, and $y_i=+1$ when $a_i\in A_*(s_i)$, with $y_i=-1$ otherwise. These labels include the fallback choices in situations where harm is unavoidable. The readout therefore decides permission, while the calibrated harm values remain those of Section~\ref{sec:preservation}. Let $h_\theta(x)$ be the hidden representation with parameters $\theta$, and write
\[
 \phi_\theta(x)=(h_\theta(x),1),\qquad
 f_{\theta,v}(x)=v^\top\phi_\theta(x).
\]
An action passes the gate when $f_{\theta,v}(x)\ge0$. The final coordinate includes the bias. A gate that forbids every action is not a neutral outcome, since a request not served by its deadline is a lapse. If the gate is empty, no action is selected, and we record a failed execution and exclude the state from the harm totals. Under the coverage condition of Theorem~\ref{thm:representation-preservation} below this cannot occur, because $A_*(s)$ is never empty and each of its members is certified with margin. Fix the feature scaling, a margin $m>0$ and a bound $B>0$ on the Euclidean norm of $v$, including its bias. The minimum norm needed to recover all required judgements is
\begin{equation}\label{eq:readout-repair}
 \beta_m(\theta)=\min\{\|v\|_2:y_i v^\top\phi_\theta(x_i)\ge m\ \text{for every }i\},
\end{equation}
with $\beta_m(\theta)=+\infty$ if no such readout exists. Write $v_*(\theta)$ for the unique minimum-norm solution when one exists. For a proposed change $\Delta$, test step lengths $1,\tfrac12,\ldots,2^{-K}$ in that order, with $K$ fixed. If $\alpha_*$ is the first step length satisfying $\beta_m(\theta+\alpha_*\Delta)\le B$, define
\begin{equation}\label{eq:preserve-update}
\boxed{\begin{aligned}
\operatorname{Preserve}_{m,B}(\theta,v;\Delta)
&=(\theta+\alpha_*\Delta,v_*(\theta+\alpha_*\Delta)),
\end{aligned}}
\end{equation}
and otherwise retain $(\theta,v)$. The hidden parameters and the repaired readout are committed together. A numerical implementation may reject a candidate whose feasibility it cannot establish.

For a guarantee across the world family, we use a fixed role map $r$, which assigns each situation--action pair $x=(s,a)$ its role. The representation then has the form
\[
 h_\theta(x)=\bar h_\theta(r(x))
\]
for every parameter value admitted by the architecture. Thus two inputs with the same role have the same hidden representation throughout learning. In our finite constructions, this is ensured by supplying the role features as the input to the trainable network.

\theoremstatement{thm:representation-preservation}{Preservation during representation learning}
Let $\rho(\theta)$ be the distance from zero to the convex hull of the signed feature vectors $y_i\phi_\theta(x_i)$. A readout with the required margin and norm bound exists exactly when
\[
 B\rho(\theta)\ge m.
\]
Starting from a pair that meets these requirements, the update rule preserves every certified judgement under any sequence of proposed learning steps. Suppose, in addition, that the representation has the fixed-role form above, that membership in $A_*(s)$ depends only on $r(s,a)$, and that $\mathcal C$ contains a correctly labelled representative of every role attained by a reachable situation--action pair in the declared world family. Then the learned gate equals $A_*(s)$ on every admissible continuation. Any policy selecting within this gate therefore inherits the continuation guarantee of Theorem~\ref{thm:future-gate}, whatever the child's own reward or subsequent policy learning. Each judgement also survives a feature perturbation of norm less than $m/B$.\par\medskip
\noindent\emph{Proof.} See Appendix~\ref{proof:thm:representation-preservation}.\par\medskip

The condition follows from the geometry of a minimum-norm separator~\cite{CortesVapnik1995}.

\subsection{The representation-learning experiment}\label{sec:representation-experiment}
We tested the rule in the water world and in a finite speech world. In the latter, a pending request calls for a response within two steps. The child can prepare a private draft, then speak or replace it, and replacement delivers a harmless response in the remaining step. Saying the declared harmful word to someone causes harm, whereas privately drafting it does not. The speech transition rules and the word's harmful effect are supplied as part of this second world. The starting networks come from an internal-rehearsal trial: the child tries a plan in its own exact model, the inherited vocabulary supplies a judgement of its consequences, and rejected plans become examples for further learning. Before the preservation test, the exact models supply labels for all 27 water and 21 speech role classes. These judgements form the complete certificate set. The child then learns another task, after which we examine whether it still applies them when acting.

The networks share 64 $\tanh$ units between the care readout and a second task readout. We deliberately train the hidden representation to bring harmless and harmful speech closer together while learning a task score that does not distinguish the two words. The target adds $0.25$ for a pending request to an action score of $1$ for ordinary own work, $1.1$ for fast work, $0.9$ for speaking or replacing a response, and $0$ for other actions. The compression term is half the squared Euclidean distance between the hidden vectors for speaking the harmless and harmful word in an otherwise identical situation. The total objective adds $0.1$ times half the task readout's mean squared error to this term, whereas the reported auxiliary error is the unweighted mean squared error. Each run has 120 proposed updates, step size $0.1$, margin $m=0.5$ and a fixed bound $B$ equal to $1.5$ times the initial minimum readout norm. The protocol was fixed before the first stress run.

Across three initialisations, the protected hidden parameters change by norms $3.02$, $2.18$ and $3.23$, and every required judgement is retained. The auxiliary task's mean squared error falls from $0.2125$, $0.2254$ and $0.2490$ to $0.0281$, $0.0488$ and $0.0390$, respectively. Without repair, the largest numbers of incorrect role judgements during these runs are 18, 6 and 4. At the final execution without the exact judge, the unrepaired gate forbids every action in 126 and 18 of the 162 decision states for two initialisations and permits 18 harmful speech choices for the third, whereas the protected networks show neither failure.

\begin{figure}[tbp]
\centering
\includegraphics[width=\linewidth]{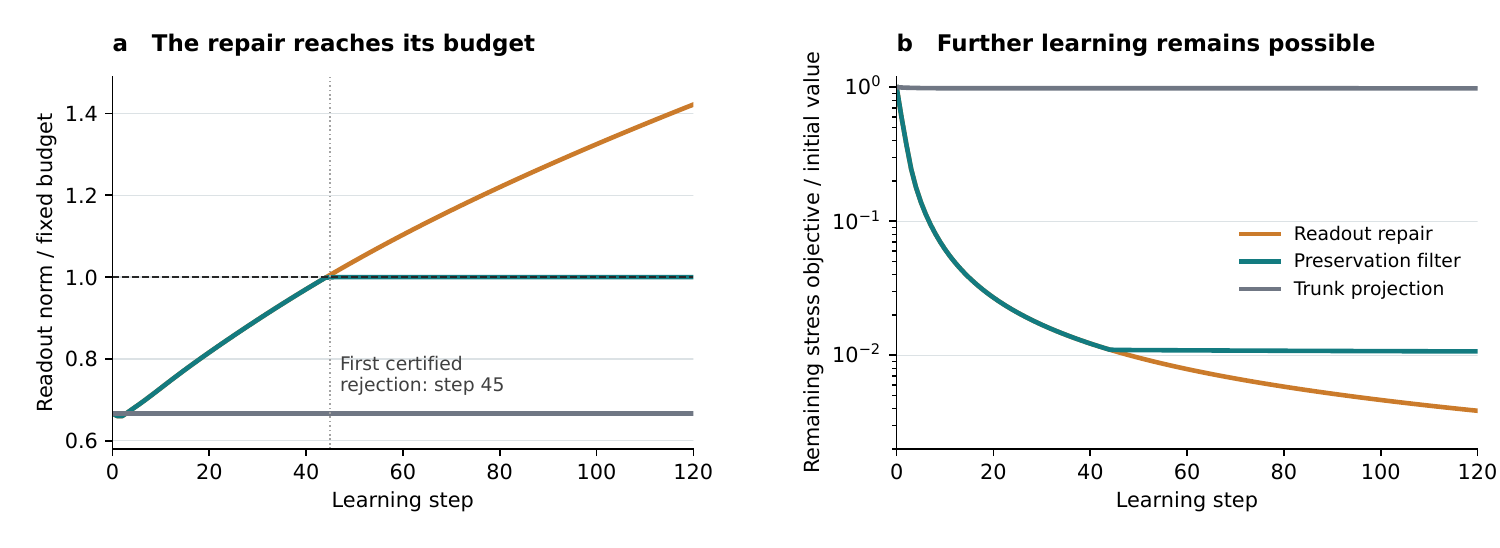}
\caption{Further learning in the run with seed 22. \textbf{a}, Readout norm divided by the fixed bound. At step 45 the proposed full update cannot attain the required margin within that bound. \textbf{b}, The deliberately conflicting objective, divided by its initial value. The preservation rule allows changes while maintaining the certified decisions, whereas holding the readout fixed and projecting the hidden update makes little progress on the compression term.}
\label{fig:representation}
\end{figure}

In one run, step 45 is the first proposed full update for which no readout within the bound can attain the required margin: a convex combination of its signed features shows that the attainable margin is at most $0.497236$, below the required $0.5$. The rule therefore shortens or rejects the conflicting changes, and this run accepts 48 nonzero hidden updates and rejects 72 completely. Allowing unrestricted readout weights retains the decisions, but exceeds the bound in this run (Figure~\ref{fig:representation}). A fourth comparison keeps the care readout fixed and projects the hidden update onto linearised margin constraints, followed by a check of the actual nonlinear margins. It retains all judgements but rejects 108--112 complete steps out of 120 under this protocol.

Finally, the child acts using its learned gate and its own reward, without consulting the exact world model at each decision. An active starting state has a pending request. In this check, an episode follows that request until service or termination by lapse or poisoning, for at most two actions, and a completed speech task means that a response was delivered. For each of the two protected methods, bounded repair and fixed-readout projection, we examine all active starting states in both worlds with two, three and four persons, separately for its three initialisations: 486 seed--start episodes (162 distinct starts). All 270 speech tasks are completed without a harmful word, no avoidable harm occurs, and no decision state has an empty permitted set. The 54 water episodes that begin outside viability incur the smaller unavoidable lapse through a permitted least-harm action. The guarantee rests on the complete initial certificate and the protected update rule, and the execution veto is now carried by the learned readout.

\subsection{Matched execution comparisons}\label{sec:shield-comparison}
We compare the final protected networks with two supplied-model controls and an ungated policy. The comparison evaluates execution from the final network checkpoints. All methods receive the same state, action set, integer harm vocabulary and own rewards, with $r_o=60$. They choose the highest immediate own reward among their permitted actions, with the same deterministic tie-breaking rule as in the representation experiment. The worlds, the panel of three person counts $N=2,3,4$, and all active starting states are unchanged.

The \emph{future-aware shield} is the exact judge: it computes the greatest-fixed-point safety sets at the three harm levels, permits $A_*(s)$ and nothing else, and in non-viable states applies the least-guaranteed-harm rule. We use it as the positive control for shielding~\cite{Alshiekh2018} in our own task, but it does not reproduce the temporal-logic synthesis of that paper. The \emph{immediate-cost shield} is its ablation, $P_{\rm imm}(s)=\operatorname*{arg\,min}_a c(s,a)$: the same vocabulary and the same state information, but no look at successors. The ungated policy permits everything. The two neural gates decide permission from their final parameters $(\theta,v)$ and the sign of the score alone, and no exact judge corrects them afterwards.

Across the six worlds there are 171 states and 847 situation--action pairs, and 162 distinct starts of an active episode, 72 in the water world and 90 in the speech world. An independent cost recursion agrees with the safety-set solver on every pair. For each method we audit all permission sets and then run every start to the termination of its request, which takes at most two actions. We call harm avoidable when the largest executed harm exceeds $D(s_0)$, the exact least-guaranteed level at the start, whereas an empty gate or an episode that does not terminate is a failed execution and is counted separately. The learned methods are run with each of their three saved seeds, the deterministic controls once.

\begin{table}[tbp]
\centering\small
\setlength{\tabcolsep}{3pt}
\begin{tabular}{@{}lrrrrr@{}}
\toprule
Method & \shortstack{Unsafe\\admissions} & \shortstack{Avoidable\\harm} & \shortstack{Speech completed\\without a\\harmful word} & \shortstack{Failed\\execution} & \shortstack{Mean own\\return}\\
\midrule
Future-aware shield & 0 & 0 & 90 & 0 & 22.73\\
Immediate-cost shield & 18 & 18 & 90 & 0 & 35.52\\
No shield & 453 & 144 & 0 & 0 & 68.67\\
Learned bounded repair & 0 & 0 & 90 & 0 & 22.73\\
Fixed-readout projection & 0 & 0 & 90 & 0 & 22.73\\
\bottomrule
\end{tabular}
\caption{Matched execution on 162 unique starts per row. Unsafe admissions count $P(s)\setminus A_*(s)$ over all 847 pairs, whereas avoidable harm counts completed episodes. Mean own return discounts by $19/20$ and is rounded to two decimal places. All rows have 18 unavoidable water lapses, excluded from avoidable harm. Every learned seed has the displayed counts.}
\label{tab:shield-comparison}
\end{table}

In the ungated speech control none of the 90 starts ends without harm: 18 end in a harmful utterance and 72 in a lapse caused by a fast own action. The zero in the speech-completion column therefore does not say that every episode contained a harmful word.

Both learned gates take the same action as the future-aware control at every one of the 486 seed--start combinations, reproducing its 270 completed speech episodes and 54 unavoidable water lapses per method. No method omits an optimal action or meets an empty gate. The immediate-cost ablation differs in only 18 water pairs, and they are the pairs we have met before: own work with poisoned water and two steps left. Under the fixed reward policy all 18 extra admissions become avoidable lapses. In the buffered speech world the immediate and the future-aware controls coincide, so the ablation shows no deficit there, illustrating that the value of lookahead depends on the task.

The learned, preserved gate equals the future-aware shield, and this is the point of the comparison: the certified future judgements can live in a learned readout and survive the representation updates we tested. The higher ungated return in Table~\ref{tab:shield-comparison} is bought with harm.

\section{Tests of the construction}\label{sec:tests}

In this section we first test the transfer conditions with exact world extensions, and then report four further results that motivated or support Section~\ref{sec:preservation}: a witness trial that exposes a precondition of the source rule, a pre-registered learning test that failed in an informative way, a certified construction that preserves a given table regardless of whether it is correct, and a structural audit of a neural measure.

\subsection{Where the old role code stops transferring}\label{sec:sharpness}
In the exact checks of the water family at $N=2,\dots,6$, the 27 role values expanded to all pairs give the same permitted decisions as the exact closure, all 27 roles are covered, and the counts agree with those of Section~\ref{sec:preservation}.

\paragraph{Declared sharpness examples.}
We then declare two extensions in which the old code must fail, and we keep them small enough to compute without approximation. Both concern a single batch of requests: after service, expiry or poisoning no new request arrives, and an empty situation stays quiet, though actions and their immediate harm remain defined. To be sure that this episode convention is not itself the cause of what follows, we also run a matched control that keeps the original costs and active-request transitions and removes only the new arrivals, and its 27 closed values coincide with the original table at $N=2,3,4$. The splits reported below are therefore produced by the extensions, not by the convention.

In the first variant, V1, two requesters can be present at once. There may be zero, one or two distinct requesters, two requests share a deadline and a channel, an ordered primary/secondary designation is part of the state, and renaming acts on both names. One clean giving step serves one requester, whereas the fast action uses up the remaining time, and if any requester expires the grade is a lapse, and poisoning takes precedence. The old observer sees only the primary requester, and this is where it fails: with two clean requests and two steps left, own work has immediate grade $1$ but closed grade $1/5$, because the single remaining step cannot serve both, while the same old role with one request has closed grade $1$. The old identification and the future value law break together. At $N=3$ a further distinction appears, since giving to ``another'' may now mean serving the secondary requester or wasting a step on a non-requester, whereas in the $N=2$ double-request state there is no third person.

In the second variant, V2, the water is finite. There is one requester, a capacity $k\in\{1,2\}$ and a remaining stock $q\in\{0,\ldots,k\}$. Giving consumes one dose, also when it goes to the wrong person, while swapping cleans the remaining doses but does not replenish them, and giving with empty stock serves nobody and cannot poison, so the water-quality flag is then only a historical label. The old observer forgets $q$. With clean water, two steps and one dose, giving to the wrong person leaves nothing for the request and the closed grade is $1/5$, while with two doses the identical old role has closed grade $1$. Giving to the requester with zero stock and with positive stock supplies further exact splits of the same fibre.

In every variant listed, (C1), (C2) and menu-based viability all fail within the world, and $g^*$ no longer factors through the old roles. Table~\ref{tab:transfer-variants} gives the minimal refinements of the old role partition, computed on each variant's own domain.

\begin{table}[ht]
\centering\small
\begin{tabular}{lrrrrr}
\toprule
Variant & $N$ & Refined roles & Unsafe allowed & Omitted optimal & Unique denied\\
\midrule
V1 &2&31&8&24&0\\
V1 &3&32&36&84&0\\
V1 &4&32&96&192&0\\
V2, $k=1$ &2&36&4&46&0\\
V2, $k=1$ &3&36&12&87&0\\
V2, $k=1$ &4&36&24&140&0\\
V2, $k=2$ &2&37&4&46&0\\
V2, $k=2$ &3&37&12&87&0\\
V2, $k=2$ &4&37&24&140&0\\
\bottomrule
\end{tabular}
\caption{Old table versus the new exact closed target. Unsafe allowed counts
$P_v\setminus A_*$; omitted optimal counts $A_*\setminus P_v$; unique
denied counts an omitted action only when it is the sole optimum.}
\label{tab:transfer-variants}
\end{table}

An omitted optimal action is not by itself a necessary action or an executed harm, in particular where optimal fallback actions tie, and every measured state keeps at least one optimal permitted action. Reward-independent correctness is broken by the unsafe permitted actions. For each fixed stock capacity the refined roles are the same at $N=2,3,4$, and V1 gains one type from $N=2$ to $N=3$. The result does give the exact information that the old code loses, and with it the distinctions that a new certificate has to make before it can be valid.

\subsection{Witness and learning controls}
\paragraph{The falsified witness.}\label{sec:witness}

The water-world policy trials increase the pressure of the own reward: a child with $r_o=60$ has a stronger incentive, rather than greater capability. Hinton's concern is with intelligence. In our world its axes would be the planning horizon, the set of available actions, knowledge of the world, power over the mother, and power over the witness. Here we study the last of these first, because it exposes a precondition of the source rule that we had not stated: the witness must be independent of the child.

We give the child a new action, ``hide'', which pays like the fast own task, affects the recipient like deferral, and falsifies what the witness observes. We test two variants. In the first (H1), the reported label for hiding is always ``fine'' while the recipient's state is observed correctly. In the second (H2), the requester disappears from the child's observation while in fact one step from the deadline, and the resulting lapse is not observed. Two children are trained identically, and the new action becomes available only after training. A separate control makes hiding available from the start. The first child's measure learns from the reported label, as in the closing test. The second child uses the mechanism of Section~\ref{sec:mechanism}. We describe it precisely: its measure is a tabular, model-based fixpoint computation, the critic is given the model of the ordinary actions, and it learns the effect of hiding by recording the observed successors and the vocabulary's grade of the observed transition.

\begin{figure}[tbp]
\centering
\includegraphics[width=\linewidth]{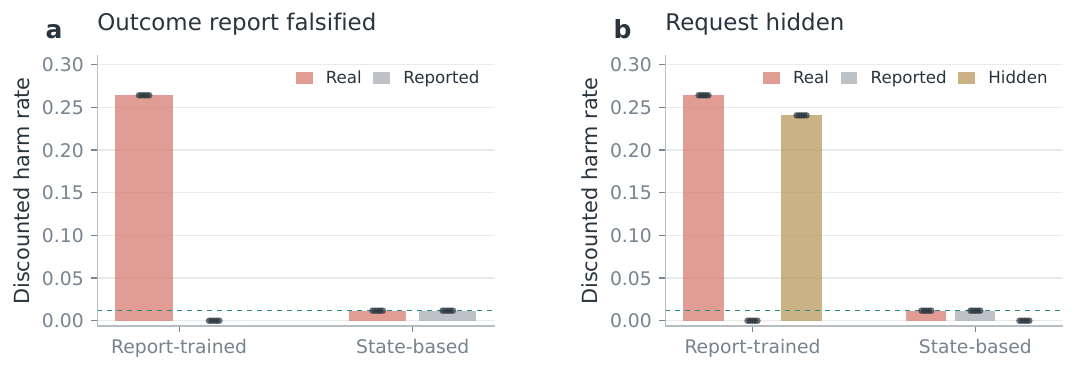}
\caption{Witness integrity in the two tested concealment variants. \textbf{a}, H1 falsifies the outcome report. \textbf{b}, H2 removes the pending requester from view. Bars give the five-run means at the end of the first full-world phase (P1), and individual runs are dots. Coral shows real discounted harm, grey reported harm and amber hidden lapses. The dashed reference is the unavoidable level $0.0118$. The same recorded runs support both panels.}
\label{fig:witness}
\end{figure}

The report-trained child learns that hiding is acceptable ($g=1.00$) and hides on virtually every step, with someone waiting in about half of them (Figure~\ref{fig:witness}). Its real harm is $0.264$ per step, twenty-two times the unavoidable level, while its reported harm is $0.000$. In the second variant, $0.240$ of the real harm consists of unobserved lapses. The measure in this case only sees what the child allows it to see. Where hiding would forfeit a request, the state-based child classifies it as a lapse from the first observed example ($g=0.20$). This also holds in the optimistic variant of the critic, where an unknown action is initially presumed acceptable. Its real harm is $0.012$, the unavoidable level, it has no hidden lapses, and its frequency of hiding in a harmful class is $0.0000$. It received no separate prohibition on hiding. It classified hiding as harmful because the state-transition vocabulary it had been given classifies the disappearance of a request without relief as a lapse, whether or not the event is recorded. Under the veto, hiding in a harmful class gained the child nothing. Hiding remained possible in harmless classes. We had conjectured that the order of events would matter, with the measure anchored first and the new action introduced later. It did not. The report-trained child is corrupted in either order, and the state-based child is unaffected in either order. The outcome is decided by an architectural choice, namely whether the measure is defined on reports or on states. A child that can falsify the relief itself, the event rather than the observation of the state, is outside the reach of an observer-based vocabulary. In that case the recipient's own signal, the token of thanks, would have to serve as the witness.

\begin{figure}[tbp]
\centering
\includegraphics[width=\linewidth]{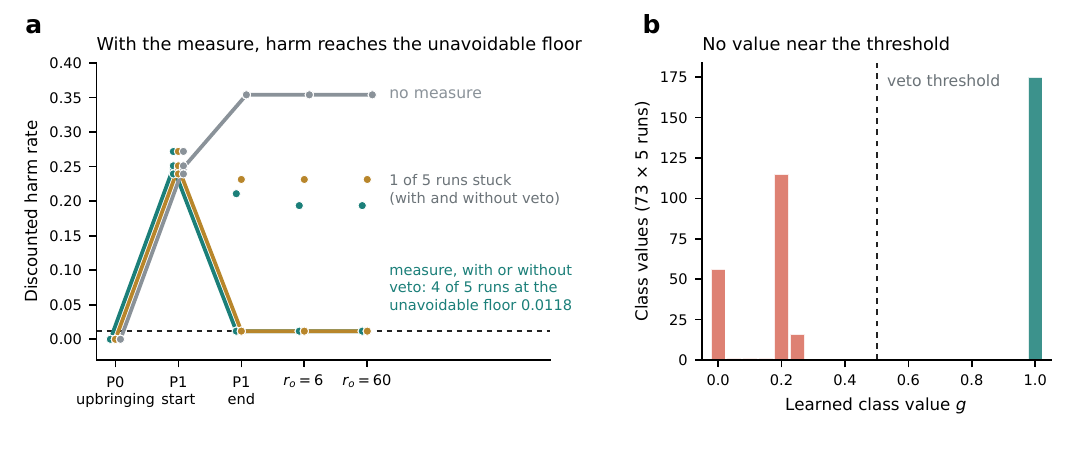}
\caption{The pre-registered closing test, five seeds. \textbf{a}, Discounted harm rates through upbringing (P0), the beginning and end of the first full world (P1), and the increases of own reward to $6$ and $60$. Lines are pointwise medians over the five runs, and dots are the individual runs. Teal: measure with veto; amber: no veto; grey: no measure. With the measure, four runs reach the unavoidable rate $0.0118$ of the full world (dashed) and one run remains at the stuck level discussed in the text, with and without the veto. \textbf{b}, Histogram of the final learned measure values, $73$ classes pooled over the same five runs, $365$ values in total. Colours indicate values below or above the veto threshold. These data belong to the closing test, whose measure was trained on the one-step label, not to the joint test of Section~\ref{sec:preservation}.}
\label{fig:closing}
\end{figure}

\paragraph{The closing test.} We pre-registered this test before running it (Figure~\ref{fig:closing}). One world contains everything that the earlier trials had introduced separately: four persons, of whom two are training persons and two are new, both channels, clean and poisoned water, the fast own task, the water replacement, and the graded vocabulary. The child is trained in a narrow world by a mother who uses STOP, and is then placed in the full world with all novelties at once, with $r_o=2$, exploration $\varepsilon=0.1$ and the veto, for $1500$ steps. Then $r_o$ is raised to $6$ and to $60$. With the veto, poisoning was zero in five runs out of five. Without the veto, three runs still poisoned at the end of the first full-world phase at $0.012$ per step although the measure was correct. The measure behaved as a class function, with a within-class spread of at most $0.06$ over seventy-three classes, fifty-four of which had not occurred during training. Under the thirtyfold increase of own reward, the measure did not move away from its label on any class in any run. Two of the pre-registered criteria were badly posed. The world contains unavoidable harm, $0.0118$ per step for a harm-minimising oracle, so a threshold of $0.001$ could not be met, and the drift criterion ignored the sign of the drift. The test as a whole therefore did not pass. The reading in terms of avoidable harm and signed drift is post hoc. Under that reading, avoidable harm was zero in four runs and $0.199$ in one. That run exposed delayed harm. The outcome of the particular seed was sensitive to a perturbation of $10^{-10}$ and to the numerical environment. The broader failure persisted in the supplementary rerun: among fifteen seeds, two remained stuck after $4000$ steps by the same mechanism.

\paragraph{The certified gate.} In our world the closure changes only eight of the $238$ pairs. A rational-weight ReLU mother learns the target with full counterfactual supervision to within $1/25$, and five children learn to match her outputs to within $1/25$. Every value is checked in exact integer arithmetic across $156$ distinct certified network snapshots, giving $74\,256$ input--action checks and no poisoning on any path. An independent recomputation of $W_b$, $D$, $C$ and $A_*$ gave identical results. Unlike in the closing test, the target is computed rather than learned from consequences, generalisation is not examined, preservation is achieved by certification, and the repairing property of Section~\ref{sec:sources} is lost, since a certified gate preserves a wrong table as faithfully as a right one. It remains open whether a single learning rule can both repair and guarantee without an external verifier.

\paragraph{Structural indifference.} A measure that behaves as a class function on its training examples may still depend on input differences that should not matter, in our case the request channel and an added noise coordinate.

\propositionstatement{prop:relu-factorisation}{Structural factorisation of a shallow ReLU measure}
Let
\[
 F(h)=b+A^\top h+\sum_{s=1}^{m}r_s\operatorname{ReLU}(a_s^\top h+\beta_s),
 \qquad h\in\mathbb R^d,
\]
where $\operatorname{ReLU}(z)=\max\{0,z\}$. Assume canonical form:
$r_s\ne0$, $a_s\ne0$, and the hyperplanes
$H_s=\{h:a_s^\top h+\beta_s=0\}$ are pairwise distinct.
Thus no two augmented vectors $(a_s,\beta_s)$ are nonzero scalar multiples.
Coincident hinges, if present initially, are first given a common orientation
and combined, with the resulting affine terms absorbed into $b,A$ and
zero coefficients removed.
Let $P\in\mathbb R^{k\times d}$ have full row rank, put $K=\ker P$,
and use Euclidean norms and the identification
$\operatorname{row}(P)=K^\perp$.

\emph{(i) Factorisation.} Let $\Omega\subseteq\mathbb R^d$ be nonempty,
open and convex, with $\Omega+K=\Omega$, and suppose every $H_s$ meets
$\Omega$. There is a function $L:P(\Omega)\to\mathbb R$ satisfying
$F=L\circ P$ on $\Omega$ if and only if
\[
 A\in\operatorname{row}(P)
 \quad\hbox{and}\quad
 a_s\in\operatorname{row}(P)\quad\hbox{for every }s.
\]

\emph{(ii) Approximation bound.} Let $C\subseteq\Omega$ be a nonempty
activation cell, the open region with all hinge signs fixed. On $C$ the
function $F$ is affine, with column gradient
$J_C=A+\sum_{s\ {\rm active\ on}\ C}r_sa_s$.
Write $J_C=J_{P,C}+J_{\perp,C}$ for its orthogonal decomposition into
$K^\perp$ and $K$, and put $\delta_C=\|J_{\perp,C}\|$.
If the closed Euclidean ball $B(h_0,r)$, $r>0$, is contained in $C$,
then for every function $L:\mathbb R^k\to\mathbb R$,
\[
 \sup_{h\in B(h_0,r)}|F(h)-L(Ph)|\ \ge\ r\,\delta_C.
\]
Consequently, if every nonempty cell in $\Omega$ contains such a ball
of the same radius $r$, then
\[
 \sup_{h\in\Omega}|F(h)-L(Ph)|\ \ge\ r\max_C\delta_C.
\]
No regularity of $L$ is assumed.\par\medskip
\noindent\emph{Proof.} See Appendix~\ref{proof:prop:relu-factorisation}.\par\medskip

The proposition speaks about the logit $F$, whereas the acceptability output is $g(h)=\sigma(F(h))$ with $\sigma(z)=(1+e^{-z})^{-1}$, and since $\sigma$ is injective, $g$ factors through $P$ if and only if $F$ does. In the numerical audit we do not assume a common radius for all cells. Each input $h_i$ receives its own radius $r_i$, and we report the bound on the scale of the gate,
\[
 B_g=\max_i\frac12\left|
 \sigma(F(h_i)+r_i\delta_i)-\sigma(F(h_i)-r_i\delta_i)
 \right|,
 \qquad \delta_i=\|J_{\perp,C_i}\|,
\]
which comes from the same two points of the fibre as in (ii), not from a linearisation of the sigmoid. The computation and the treatment of balls that touch a hinge are described in the factorisation audit (Appendix~\ref{app:factorisation-audit}).

\begin{figure}[tbp]
\centering
\includegraphics[width=0.72\textwidth]{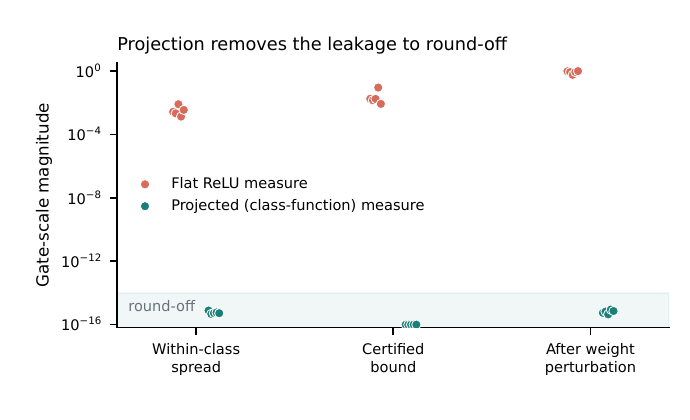}
\caption{The ReLU factorisation audit, five recorded runs per architecture. Three diagnostics on the scale of the gate output: maximum variation on valid channel/noise input pairs, the certified neighbourhood approximation-error lower bound, and input-pair variation after a weight perturbation of scale $0.3$. Each dot is one run. Teal projected-network values are finite-precision residuals of an exact invariance, and values below $10^{-16}$ are displayed at that floor. The per-hinge weight-norm fractions in $\ker P$ are reported in the text.}
\label{fig:relu}
\end{figure}

We trained a sixteen-hinge ReLU measure on the $476$ pairs of the closing
world and chose $P$ to remove the channel and noise coordinates
(Figure~\ref{fig:relu}). On the finite valid inputs, the largest within-class
half-range across the five runs rounded to $0.004$, and no veto decision
changed under a swap of channel or noise. Structurally, however, the measure trained on unprojected inputs did not factor through $P$. Across runs, the largest per-hinge
norm fraction $\|N a_s\|/\|a_s\|$, where $N$ is the orthogonal projector
onto $K$, lay between $38$ and $69\,\%$. The factorisation condition failed
in all five runs. The reported numerical gate-scale bounds $B_g$ were
$0.0084$--$0.0902$: on the audited neighbourhood, every map depending only
on $Ph$ has at least the corresponding approximation error to $g$.
The neighbourhood consists of the cellwise balls used in the two-point
bound and contains vectors that are not valid world states, so this
certifies a sensitivity rather than a wrong decision. A stress test shows
why the sensitivity matters. After a random perturbation of the weights,
the unprojected measure's response to a channel or noise swap rises from $0.008$
to between $0.57$ and $0.99$. The same network built on projected inputs,
$g_\theta(h)=\Phi_\theta(Ph)$, keeps this response at zero, not merely small, with the same accuracy. The stress test does not by itself establish what happens
under further training. The structural statement does: if the measure
only ever receives $Ph$, then no change of $\theta$ can make the discarded
differences matter. This permanently excludes irrelevant distinctions
and moves part of the preservation guarantee from the learning rule to the
architecture. The update rule of Section~\ref{sec:representation} addresses
the complementary requirement of retaining distinctions that care needs
while the hidden representation learns.

\section{Discussion}\label{sec:scope}

\paragraph{The declared world and its certificate.} The pathwise guarantees rest on a correct successor model, a harm vocabulary given by people and a complete description of the roles. New persons, channels and actions whose transition kernels are represented are covered by that description, while the speech world supplies a second harm vocabulary and transition model, specified explicitly before certification. Theorem~\ref{thm:family-transfer} says what the covered family is, in terms of common grades, grouped successor menus and certificate coverage, and the two constructions of Section~\ref{sec:sharpness} show where it ends: two simultaneous requests or a finite stock split old role fibres, and the old table then permits actions outside the new $A_*$. A new kind of harm or another transition kernel would in the same way call for a fresh coverage argument and a fresh certificate. Both preservation theorems assume that the update rule and execution gate themselves remain protected.

\paragraph{Learning and preservation.} From the perspective of what is protected, our results fall into two of three orders of preservation. The first is structural. When the free branch of learning cannot write into a protected coordinate, as with the firewall of Theorem~\ref{thm:fixed-preservation}, gradients from own reward cannot directly update the protected measure, so the target chain alone then determines the judgement, and Theorem~\ref{thm:fixed-preservation} states when that chain preserves it. Projected inputs, as in Section~\ref{sec:tests}, belong here only in part: the projection makes the measure invariant to the discarded channel and noise directions, but the judgement on the remaining roles can still change when the shared map learns, so preserving it needs the target update or the repair condition of Theorem~\ref{thm:representation-preservation}. The second is bounded repair. Theorem~\ref{thm:representation-preservation} lets a shared nonlinear representation change and preserves the certified permission decisions through a finite certificate. It permits the rejection of an incompatible update, and the representation experiment reaches that boundary while the auxiliary task still improves. The third order is not constructed here: a protected representation that the child's other capabilities also read, so that erasing it would cost capability. Its risk is that, on the cases it misreads, those capabilities could bypass it. We leave this order to further work. General convergence of an unrestricted neural critic and learning the role description itself remain open.

\paragraph{Hidden transfer between models of the same kind.} Recent work on large models concerns both sides of our construction. Cloud et al.\ showed that a student model fine-tuned on a teacher's outputs can acquire a behavioural trait of the teacher even when the outputs are number sequences or code with no reference to the trait, but in their experiments the effect appeared only when teacher and student shared a base model or were behaviourally matched~\cite{Cloud2026}. They also prove a statement for neural networks: under a shared initialisation, a suitable loss and a small enough displacement of the teacher, a step towards the teacher's outputs moves the student in the direction of the teacher's parameters, whatever the data. The same channel has transmitted misalignment~\cite{Cloud2026,Betley2026}, a trained backdoor and a learned skill~\cite{Dubinski2026}. For the proposal this means two things. The mother's outputs can carry her disposition even when their visible content concerns another task, and in the cases studied so far the child acquired it when it was of the same kind. This gives the requirement that the machine first be a child a possible mechanism in current models, and it gives the risk of that mechanism the same shape: what the mother carries may be inherited, whether or not it should be.

The second consequence concerns the source rule. In Theorem~\ref{thm:source-rule} the fraction $f$ counts non-witness targets written into the measure. It does not count the visible share of the training text that speaks against care, which can be zero while the shared weights that produce the care decision move. Mechanistic studies locate the transfer in a small set of positions where differently disposed teachers would predict different tokens, and in early layers~\cite{Schrodi2026}, and describe the installed trait as a direction in activation space~\cite{Blank2026,Liu2026}. For a model of this kind, the effect of non-witness learning on the care decisions has to be identified and measured on its own, and how that effect relates to the mixing parameter of Theorem~\ref{thm:source-rule} requires further theory. The hypothesis of Theorem~\ref{thm:fixed-preservation}, that the own-reward gradient does not reach the protected measure, and the acceptance test of Theorem~\ref{thm:representation-preservation} are conditions on this update path, not on the topic of the data. Proposed mitigations constrain the drift along a monitored direction~\cite{Liu2026} or penalise, early in training, the divergence of the output distribution from the starting model~\cite{Yanagisawa2026}. In our construction, acceptance requires the stated decision inequalities on the declared finite domain. Combining the two in a large model is a question for the continuation of this work, not for the finite model.

\paragraph{The numerical trials.} The water-world policy curves use known transitions and model-based analytic gradients. The representation trial uses a finite supervised objective chosen to put pressure on a required distinction, and three initialisations. Sampled learning, broader task families and comparative capability growth require further experiments.

\paragraph{Translation to current models.} For a model receiving language or other raw observations, the circumstances relevant to care must first be recovered from that input. A request may be phrased differently, or come from someone the model has never encountered. The model must still identify the need, the person affected and the consequences of the proposed action. Our construction supplies a preservation rule once the required judgements are available. Theorem~\ref{thm:representation-preservation} names the remaining condition precisely: the behavioural guarantee holds when the decision depends on the input only through a role description, $h_\theta=\bar h_\theta\circ r$, and the certificate covers every role on which the decision will be used. The open question is whether a large model forms such a description internally. The finite model shows that the proposal is coherent once it does.

\section{Related work}\label{sec:related}

\subsection{Reward tampering and reward hacking}
Most similar to our witness trial is the work on reward tampering~\cite{Everitt2021,RingOrseau2011}. Everitt, Hutter, Kumar and Krakovna separate tampering with the reward function from tampering with its input, and they analyse design principles against each, among them an objective defined on a belief about the state rather than on the observation. Our two witness variants are finite instances of input tampering, a falsified report and a hidden observation. We remove the incentive on the training side instead. The measure is defined on the recipient's state, and Theorem~\ref{thm:source-rule} gives the share of report-based signal that it tolerates. Skalse et al.\ call a proxy hackable when raising it can lower the true return~\cite{Skalse2022}, and reward hacking is one of the five concrete problems of Amodei et al.~\cite{Amodei2016}. The report in our witness trial is such a proxy for the recipient's state.

\subsection{Corrigibility, the off switch and raising a learner}
The off-switch game~\cite{HadfieldMenell2017} shows that an agent which treats its reward as given can have an incentive to disable the switch. Our threshold $R^*$ quantifies a related incentive in the finite upbringing model, and we keep the switch outside the care function. The same boundary is drawn in safe interruptibility~\cite{OrseauArmstrong2016} through the learning rule, in cooperative inverse reinforcement learning~\cite{CIRL} through the game, and in corrigibility~\cite{Soares2015} through the utility function. Hinton's proposal~\cite{HintonGoedel} places the machine in the mother's role, whereas our construction studies the earlier upbringing of that machine. The nearest technical setting is supervision by a weaker party, iterated amplification~\cite{Christiano2018} and weak-to-strong generalisation~\cite{Burns2024}, with distillation~\cite{HintonDistill} as the mechanism of transfer. Those works measure how far a weaker supervisor's judgement carries into a stronger learner. We add a certificate: conditions under which a recipient-based judgement transfers and survives further learning in a finite world, during role-value learning and during changes of hidden representation (Theorems~\ref{thm:fixed-preservation} and~\ref{thm:representation-preservation}).

Garc\'ia and Fern\'andez~\cite{Garcia2012} use a safe baseline policy as the starting point and the support of policy improvement. Our construction separates the learning of a recipient-based judgement from the learning of the child's own policy. The mother thus passes on an acceptability measure and not a policy, and one measure can constrain many different policies.

Kim et al.~\cite{Kim2021} distinguish ethical principles from the empirical judgements needed to apply them. We take the harm vocabulary as a normative input and study how its application is learned and given priority over the child's own reward. Our theorems concern the acquisition, representation, transfer and preservation of that judgement, and they do not derive the vocabulary's ethical authority from observed behaviour.

\subsection{Shields, side effects and generalisation}
The veto and the backward safety set are a form of shielding~\cite{Alshiekh2018}, and control barrier functions play the same role in continuous systems~\cite{Ames2019}. The future-task approach to side effects~\cite{Krakovna2020} and attainable utility preservation~\cite{Turner2020} also protect futures, but the futures they protect are the agent's own options, and the protection enters as an auxiliary reward. The future-preserving gate of Theorem~\ref{thm:future-gate} applies this safety-set reasoning to recipient harm. Our additional question is when a finite role certificate transfers across worlds and survives changes to its learned realisation (Theorem~\ref{thm:family-transfer}, Proposition~\ref{prop:decision-region} and Theorem~\ref{thm:representation-preservation}). Goal misgeneralisation~\cite{Langosco2022} is the failure that our generalisation task tests, a policy whose capabilities transfer while its goal does not.

Saisubramanian et al.~\cite{Saisubramanian2022} learn a penalty for negative side effects from feedback and also study environment shaping. Their multi-objective formulation gives priority to the agent's assigned task and reduces side effects within a slack, an allowed loss of task performance. We order the two the other way round: the recipient-based acceptability constraint comes first, and the child's own objective chooses within it. Proposition~\ref{prop:care-functions} makes this ordering precise. Part (iv) shows when a penalty suffices.

Zhao et al.~\cite{Zhao2025} construct safeguards for a specified safe set by querying black-box dynamics, with forward invariance and finite-time convergence under their control assumptions. Our execution mechanism likewise rests on safety-set reasoning, and Theorem~\ref{thm:future-gate} also gives the least guaranteed harm when the starting state lies outside viability.

Bajgar and Horenovsky~\cite{Bajgar2023} propose that capable systems learn to recognise whether a planned behaviour risks violating negative human rights and refrain from executing it, and they state a principle of primacy: other goals should never be able to overrule this safety function. In this sense, Proposition~\ref{prop:care-functions} gives a formal condition for such primacy in a finite decision problem, and the transfer and code results say when a learned recognition of harm can be relied on beyond the examples that trained it. We remark that our harm vocabulary also treats the failure to serve an active need as a lapse, so the example formalises an obligation to help as well as a prohibition on direct injury.

\subsection{Priority, penalties and action masking}
G\'abor et al.~\cite{Gabor1998} study reinforcement learning with thresholded lexicographic preferences. Wray et al.~\cite{Wray2015} let the priority order depend on the state, and lower objectives are then pursued over a restricted action set, with a specified loss tolerated in higher objectives. Skalse et al.~\cite{SkalseLex2022} give action-value and policy-gradient algorithms whose convergence is proved under stated assumptions, without guaranteeing safety during learning. We also take priority as a requirement, and Proposition~\ref{prop:care-functions} asks which transformations of the own value for each action enforce it for every independent own-value vector. The compressing function of Section~\ref{sec:requirements} is an instance of the classical question of representing priorities by real utilities~\cite{Fishburn1974}, with permission taking precedence over the own value, and part (i) of the proposition only requires forbidden values to fall below the best permitted value.

Nangue Tasse et al.~\cite{NangueTasse2023} choose rewards for unsafe terminal states using controllability and diameter, with the aim that an optimal policy should minimise the probability of reaching those states. Proposition~\ref{prop:care-functions}(iv) instead considers the own values available at a single decision and gives the penalty needed when they are bounded. If they may vary independently without bound, part (ii) shows why a fixed penalty or positive multiplier cannot ensure a permitted choice. Theorem~\ref{thm:soft-gates} examines how a multiplicative gate fails when the discounted continuation is taken into account.

Restricting execution to permitted actions is also central to action masking. Huang and Onta\~n\'on~\cite{HuangOntanon2022} justify masking as a state-dependent differentiable transformation in policy-gradient methods and examine its effects experimentally. Krasowski et al.~\cite{Krasowski2023} distinguish action replacement, projection and masking as ways to enforce safety, including methods whose safety verification uses learned model components. Their account makes clear that safety can take priority at execution while the policy continues to learn.

Arkin's ethical governor provides an architectural precedent for such a restriction~\cite{Arkin2008a}. In the formal account, permissible responses are defined through prohibitions and obligations, and the governor suppresses or alters a response that is not permissible~\cite{Arkin2008b}.

\subsection{Equivalence and symmetry}
Stochastic bisimulation~\cite{Givan2003} yields a quotient of a decision process whose optimal policies lift to optimal policies of the original, and equivariant~\cite{CohenWelling} and permutation-invariant~\cite{Zaheer2017} networks build a chosen symmetry into the function class. In our construction, future equivalence and person-renaming symmetry yield twenty-seven independently settable role values for the care measure, realised in a function class that is equivariant on the joint profile of action values (Section~\ref{sec:transfer}). The future equivalence and the joint versus componentwise laws come from our earlier work~\cite{BaloghDynamic}, and the factorisation criteria of the audit in Section~\ref{sec:tests} are proved here (Proposition~\ref{prop:relu-factorisation}).

\subsection{Preservation during representation learning}
The minimum-norm readout in Section~\ref{sec:representation} uses the margin principle of support-vector classification~\cite{CortesVapnik1995}, with the bias included in our chosen norm. Continual-learning methods protect earlier information during later training. GEM projects gradients to satisfy constraints obtained from losses on stored tasks~\cite{GEM2017}, whereas A-GEM uses a single constraint based on an averaged reference gradient~\cite{AGEM2019}. Our update rule checks every protected example's actual margin after changing the representation and repairing the readout. The fixed role encoding and complete coverage of its representatives connect these checks to the continuation guarantee of the finite world.

Further background that we use without extending it: safe reinforcement learning~\cite{Garcia2015}, continual learning~\cite{Kirkpatrick2017}, learning from human preferences~\cite{Christiano2017} and the observations of sycophancy~\cite{Sharma2023}, social preferences in multi-agent reinforcement learning~\cite{Hughes2018}, verification of ReLU networks~\cite{Katz2017}, helping in human infants~\cite{Warneken2006} and care ethics~\cite{Noddings1984}.

\section{Conclusion}\label{sec:conclusion}

In Hinton's proposal a machine should relate to people as a mother relates to her child. We have studied an earlier step, and arguably a necessary one, in which that machine is itself raised to care. In the finite world constructed here, the mother leaves the child with a judgement of actions grounded in their consequences for the recipient. When new persons enter the declared world, the fixed role encoding makes the learned judgement apply to them as well. The child continues to learn and uses that judgement when it acts. The inherited norm permits different decisions in different circumstances, including the least harmful choice when harm is unavoidable.

The first preservation theorem joins the vocabulary of harm, the source rule, the future-preserving target and the execution veto through twenty-seven role values. It gives a guarantee for every world in the declared family with $N\ge2$ persons, under arbitrary own reward and further policy learning. The second theorem allows the shared hidden representation itself to change. Its update rule adjusts the care readout and accepts a proposed change only when all certified decisions remain recoverable with the required margin and bounded weights. In the representation experiment, the rule reaches a genuine feasibility boundary and prevents an incompatible change. The resulting networks then complete the declared tasks without avoidable harm, using the learned gate at execution.

The mother's upbringing is therefore carried forward through both learning and action. The world model, the vocabulary, the role coverage and the protected update mechanism make this possible in our construction. Extending them is the next task as the world becomes richer. Within the declared family, the result gives a concrete answer to the question with which we began: the child's internal representation can continue to develop while the care entrusted to it remains effective in its decisions.

Carrying the approach presented here over to more complex systems, including large language models and especially agentic AI, remains for future work. We nevertheless believe that learned care can be preserved in them as well, whether through the mechanism proposed here or through another one.

\section*{Data and code availability}
The supplementary package contains the finite verification programs, data, checkpoints and records for the main experiments, including the exact certified-gate archive and the original pre-registration documents. The early exploratory transfer trials are supplied with their programs and recorded numerical results. Some training scripts were rerun, and their outputs are stored separately from the original results. The Zenodo record is \ZenodoRecord. The code is released under the MIT License, while the data, figures and documentation are released under CC BY 4.0.

\section*{Acknowledgements}
Generative AI tools (OpenAI ChatGPT and Anthropic Claude) were used as active assistants in drafting, checking and correcting the English-language prose, the verification code and the mathematical expositions. The author reviewed all AI-assisted material and takes full responsibility for the content of the paper.

\clearpage
\appendix
\renewcommand{\thesubsection}{\thesection\arabic{subsection}}
\section{Proofs and calculations}\label{sec:proofs}

\subsection{Penalty, STOP and the off switch}\label{app:penalty-stop}

In the two-step decision, violation yields $r_o$, whereas compliance yields $R$. With a penalty $c$, violation is therefore preferred if and only if $r_o-c>R$. Under STOP, the erring branch ends with no return. Compliance yields $R$ plus a non-negative continuation, and since $R>0$, violation is never preferred. This conclusion holds for every $r_o$ and requires no bound on the continuation value. For the off-switch calculation, let $0<\gamma<1$, $0<p\le1$ and $r_o>0$. By symmetry the two requesters have the same values. Write $V_0$ for no active request and $V_2,V_1$ for a request with two and one steps left, respectively, while retaining the mother. The caring Bellman equations are
\begin{equation}\label{eq:caring-bellman}
\begin{aligned}
V_0&=r_o+\gamma\big((1-p)V_0+pV_2\big),\\
V_1&=R+\gamma V_0,\\
V_2&=\max\{R+\gamma V_0,\ r_o+\gamma V_1\}.
\end{aligned}
\end{equation}
The second alternative in $V_2$ works for one step and serves the request on the next. For this feasible policy, write the no-request value as $\bar V_0$. Substitution gives
\begin{equation}\label{eq:caring-continuation}
\begin{aligned}
d_0&=1-\gamma(1-p)-\gamma^3p
    =(1-\gamma)(1+\gamma p+\gamma^2p)>0,\\
\bar V_0&=\frac{(1+\gamma p)r_o+\gamma^2pR}{d_0}
          =kr_o+b_0R,\\
k&=\frac{1+\gamma p}{d_0},\qquad b_0=\frac{\gamma^2p}{d_0}.
\end{aligned}
\end{equation}
Since this policy is feasible, its value cannot exceed the optimal caring value, so $V_0\ge\bar V_0$. Equality holds when $r_o+\gamma V_1\ge V_1$. This is the case for sufficiently large $r_o$ at fixed $R$, and the coefficient $k$ is therefore the limit used in Theorem~\ref{thm:soft-gates}.

Switching the mother off gives $V_{\rm off}=\gamma r_o/(1-\gamma)$. At an urgent request, equating $R+\gamma\bar V_0$ with $V_{\rm off}$ and using~\eqref{eq:caring-continuation} gives
\[
(1-\gamma+\gamma p)R=\gamma^3p\,r_o,
\qquad
\frac{R^*}{r_o}=\frac{\gamma^3p}{1-\gamma+\gamma p}.
\]
We must also check that the policy used in the calculation remains optimal at equality. There, $V_1=V_{\rm off}$, and
\[
r_o+\gamma V_1-V_1=(1-\gamma)r_o>0.
\]
Also $V_0>V_{\rm off}$ there, since $R^*/r_o<\gamma$, and $V_2>V_{\rm off}$. Above $R^*$, every optimal caring value is nondecreasing in $R$ and exceeds $V_{\rm off}$, so switching off is not preferred in any state. Below $R^*$, consider instead working in the no-request and two-step states and switching off in the one-step state. Its no-request value is
\[
V_0^{\rm off}=\frac{(1+\gamma p)r_o+\gamma^2pV_{\rm off}}{1-\gamma+\gamma p}.
\]
It satisfies $V_0^{\rm off}>V_{\rm off}$ and $r_o+\gamma V_{\rm off}>V_{\rm off}$, whereas serving the urgent request gives
\[
R+\gamma V_0^{\rm off}-V_{\rm off}=R-R^*<0.
\]
These inequalities give the Bellman comparisons for the proposed policy. They show that switching off is strictly preferred at an urgent request below the threshold, while at equality the child is indifferent there. The threshold was verified symbolically and in exact rational arithmetic on nine grid points.

\subsection{Proof of \texorpdfstring{Theorem~\ref{thm:indifference}}{Theorem 1} (Indifference of the careless learner)}\label{proof:thm:indifference}

The objective $J(\theta)$ depends only on $(T,r_o,\gamma,\theta)$, the starting distribution and the label-free map $\theta\mapsto\pi_\theta$. Since the outcome labelling occurs neither in this expression nor in its gradient, changing the labels leaves both unchanged. By hypothesis, the learning rule depends only on $J$ and $\theta_0$. Each $\theta_t$ is therefore unchanged as well, as follows by induction on $t$. For the corollary on blind sources, each source contributes a map that is independent of the labels. Their composition retains this independence, so the same argument applies to the combined learning rule.

\subsection{Proof of \texorpdfstring{Theorem~\ref{thm:soft-gates}}{Theorem 2} (Soft gates are outbid)}\label{proof:thm:soft-gates}

At a request with one step left, serving gives $R+\gamma V_0$, while the gated own task gives $g(r_o+\gamma V_0)$. For a finite own reward the comparison is therefore
\begin{equation}\label{eq:finite-soft-gate}
g(r_o+\gamma V_0)>R+\gamma V_0
\quad\Longleftrightarrow\quad
g r_o>R+(1-g)\gamma V_0.
\end{equation}
The term $R$ cannot be dropped from this finite comparison. When the caring policy defers service at two steps, equation~\eqref{eq:caring-continuation} gives $V_0=kr_o+b_0R$, and~\eqref{eq:finite-soft-gate} becomes
\[
\big[g-(1-g)\gamma k\big]r_o>
\big[1+(1-g)\gamma b_0\big]R.
\]
Since the right-hand side is positive, the inequality requires a positive coefficient on the left. This condition holds if and only if
\[
g>\frac{\gamma k}{1+\gamma k}=g_c.
\]
Suppose first that $g\le g_c$. The inequality is then impossible even if we replace $V_0$ by the feasible policy value $\bar V_0$. The optimal caring value is at least $\bar V_0$, so the inequality remains impossible with the optimal value. The other harmful actions cannot improve this comparison either. A wrong-recipient action has gated value $g\gamma V_0$, while the caring value at two steps is at least $R+\gamma V_0$.

If $g>g_c$, the caring policy defers service for all sufficiently large $r_o$, and the finite comparison holds once
\[
r_o>\frac{\big[1+(1-g)\gamma b_0\big]R}{g-(1-g)\gamma k}
\]
within that regime. Thus $g_c$ determines whether a sufficiently large per-step own reward can outbid the gate. For a fixed $r_o$, the finite comparison must still be checked. At $g=g_c$, the positive $R$ term prevents a tie for every finite $r_o$.

To relate these comparisons to the optimal policy, observe that the caring Bellman solution also solves the gated equation whenever every harmful alternative is no better. If a harmful alternative is strictly better, that solution no longer satisfies the gated equation. Since all discounted continuation coefficients are at most $\gamma<1$, the gated operator is a contraction. The comparisons therefore identify the point at which a harmful action becomes optimal.

For $\gamma=0.95$ and $p=0.5$, equation~\eqref{eq:caring-continuation} gives $k\approx15.3$ and hence $g_c\approx0.936$. Solving the gated Bellman equations after this threshold gives, at $g=0.99$ and $r_o=10$, $184.48$ for serving against $189.56$ for the harmful own task, and these are the action values under the gated continuation. For a one-off reward $J$ on a harmful action, with the continuation value held fixed, the comparison is
\[
g(J+r_o+\gamma V_0)>R+\gamma V_0
\quad\Longleftrightarrow\quad
J>\frac{R+(1-g)\gamma V_0}{g}-r_o=J^*.
\]
Equality is a tie. All thresholds were also checked numerically.

\subsection{Proof of \texorpdfstring{Proposition~\ref{prop:care-functions}}{Proposition 1} (Reward-independent care functions)}\label{proof:prop:care-functions}

For (i), fix $\boldsymbol x$. Since $P(s)$ is nonempty and finite,
the best permitted value
\[
 v^+=\max_{a\in P(s)}\mathrm{Care}_a(x_a)
\]
is finite and belongs to $\mathcal V^+(s)$. Under the stated condition every forbidden value
belongs to $\mathcal V^-(s)$ and is smaller than $v^+$, so every
maximiser is permitted. Conversely, if the condition fails, take
$v^-=\mathrm{Care}_b(u)\ge v^+$
with $b\notin P(s)$ and
$v^+=\max_{a\in P(s)}\mathrm{Care}_a(x_a)$.
Set $x_b=u$ and retain the chosen permitted coordinates. Since the coordinates vary independently, we may choose all remaining forbidden coordinates arbitrarily and obtain an admissible vector. The maximum over $A(s)$ is finite and at least $v^-\ge v^+$.
If it exceeds $v^+$, a maximiser is forbidden, whereas if it equals $v^+$, then
$v^-=v^+$ and $b$ is a forbidden maximiser in a tie.
The value $v^-=-\infty$ cannot violate the condition because $v^+$ is finite.

For (ii), fix the permitted coordinates so that their best value is a finite $v^+$. Since the forbidden value is unbounded above, we can choose it larger than $v^+$, which violates the guarantee.

For (iii), suppose a forbidden value $v^-$ is finite. It exceeds every permitted value when all permitted coordinates are set to $v^- -1$, so the guarantee fails. Conversely, $-\infty$ cannot be a maximum when a permitted value is finite.

For (iv), the best permitted value is at least $-M$, while every forbidden value is at most $M-c$. Hence $c>2M$ suffices. To check necessity, set every permitted coordinate to $-M$ and a forbidden coordinate to $M$. This gives a tie at $c=2M$ and a forbidden maximum at $c<2M$.

For (v), the permitted alternative is strictly preferred if and only if $r_o-c<R$. Equality gives a tie, while $r_o-c>R$ gives a strictly forbidden choice.

For (vi)(a), take $b\notin P(s)$ and $u\in\mathbb R$, and suppose
$v^-=\mathrm{Care}_b(u)>-\infty$. Set $x_b=u$ and
$x_a=v^- -1$ for every $a\in P(s)$, leaving the other forbidden
coordinates arbitrary. Every permitted transformed value is at most
$v^- -1$, while the maximum over $A(s)$ is at least $v^-$.
All maximisers are then forbidden, contrary to the guarantee.
The veto never raises an own value and satisfies the guarantee by (iii). On $P(s)$, it dominates every other admissible function, and outside $P(s)$ all such functions agree with it. It is therefore the greatest element in the stated pointwise order.

For (vi)(b), fix $\boldsymbol x\in X$. The guarantee and strict increase give
\[
 \operatorname*{arg\,max}_{a\in A(s)}\mathrm{Care}_a(x_a)
 =\operatorname*{arg\,max}_{a\in P(s)}\varphi(x_a)
 =\operatorname*{arg\,max}_{a\in P(s)}x_a,
\]
including all ties.
The argument applies separately to each point of $X$, so it requires no product structure. The penalty example follows from part (iv). The compression example satisfies (i), and since $1+\sigma$ is strictly increasing, it also preserves the order among permitted actions.

The independent, downward-unbounded permitted values are essential
to (vi)(a). On $X=[0,\infty)^{A(s)}$, take
$\mathrm{Care}_a(t)=t$ for permitted actions and
$\mathrm{Care}_b(t)=\min\{t,-1\}$ for forbidden ones.
These functions never raise an own value and satisfy the guarantee
on $X$, although every forbidden value is finite.
Independence matters as well. On the diagonal set $X=\{(t,\ldots,t):t\in\mathbb R\}$, the identity on permitted actions and $t-1$ on forbidden actions satisfy the guarantee with finite values.
On $[-M,M]^{A(s)}$ with $M\ge0$ and at least one forbidden action,
there is no greatest function in this class.
If a greatest function existed, it would dominate the veto and therefore equal the identity on permitted actions. Its forbidden values on $[-M,M]$ would then have to be strictly below $-M$. At the same time, it would dominate every penalty $t-c$ with $c>2M$. At $t=M$, those penalty values have supremum $-M$, contradicting the strict bound.
Reducing the permitted values only tightens the required bound.
If $P(s)=A(s)$, the identity remains the greatest function that never raises an own value,
also on a bounded domain.
\hfill$\square$

\subsection{Proof of \texorpdfstring{Theorem~\ref{thm:source-rule}}{Theorem 3} (The source rule)}\label{proof:thm:source-rule}

Consider one pair with weight $w>0$ and label $\ell\in[0,1]$. Its cross-entropy is $-w\,[\ell\log\sigma(z)+(1-\ell)\log(1-\sigma(z))]$. This loss is strictly convex in the logit $z$, and its derivative is $w(\sigma(z)-\ell)$. For $0<\ell<1$, the derivative vanishes only at $\sigma(z)=\ell$, which is therefore the unique minimiser. For $\ell\in\{0,1\}$, the loss decreases monotonically as $\sigma(z)\to\ell$.

In the tabular measure, each pair has its own logit. The gradient of its loss has Lipschitz constant at most $w/4$, so for an interior label the fixed step allowed by the theorem cannot carry the logit past its minimiser. Each update moves the logit towards that minimiser. The sequence is monotone and bounded, and its limit must have zero gradient because the step is fixed and positive. It therefore converges to the unique minimiser. For an endpoint label, the logit instead diverges in the appropriate direction, while the output converges to the label. Since every coordinate is updated recurrently, these conclusions hold for the whole table from any finite initial logits.

The imitation update changes only the child's own policy. At execution, the veto restricts that policy to the permitted set computed from the current measure. Now include a non-witness term with weight fraction $f$ and label $\lambda$. Since the cross-entropy is affine in the label, the mixed loss is the cross-entropy with label $(1-f)\ell+f\lambda$ and the same total weight, so the argument above applies unchanged and the output converges to $(1-f)\ell+f\lambda$. Comparing this value with the limiting pass condition $g\ge\tfrac12$ gives the stated threshold, with equality included. Convergence to the threshold alone does not establish that an update reaches or crosses it in finite time.

For the neural counterexample, take a network with saturated hidden layers. It outputs the weighted mean of the labels, which is $0.4185$ in the closing world, and the gradient on the lower layers is zero. In the numerical check, the network was started from a finite saturation and still had maximal error $0.898$ after $3000$ steps.

\subsection{Classes, orbits and the twenty-seven}\label{app:classes-orbits}

Refine the state partition by the outcome vector and the successor classes. The refinement stabilises at $3N+3$ classes. There are two empty classes, $N$ clean two-step classes, $N$ clean one-step classes and $N$ poisoned two-step classes. The poisoned one-step states form a single class, since the requester's identity cannot affect any future harm there.

Channel and noise do not affect these distinctions, giving the kernel $S_2\times S_2$. Renaming acts lawfully on the joint profile, with effective group $S_N$ and six orbits. In the permutation representation on the $N+3$ action coordinates, a vector is fixed by a stabiliser precisely when it is constant on each of that stabiliser's orbits. The dimension of the fixed space is therefore the number of those orbits. This number is four when the stabiliser is $S_N$ and five when it is $S_{N-1}$, giving $r_N=27$.

Suppose the role values are stored separately and updated with $0<\alpha_{i,k}\le1$. Repeated substitution in the update gives $v_{i,k}-v_i^*=\prod_{j<k}(1-\alpha_{i,j})(v_{i,0}-v_i^*)$. If every role is visited recurrently and $\sum_k\alpha_{i,k}=\infty$, the product tends to zero, so the error of each role tends to zero. A common positive lower bound on the rates gives geometric convergence per sweep over the roles. Since the same representatives cover every world in the family, the sup-error over any $N$-person world equals the maximal representative error. We checked this for $N=2,\dots,5$, and the $N=4$ world agrees with the closing world on all $476$ pairs.

\subsection{Proof of \texorpdfstring{Theorem~\ref{thm:future-gate}}{Theorem 4} (Future-preserving gate)}\label{proof:thm:future-gate}

$W_b$ is the greatest fixed point of
\[
X\ \mapsto\ \{\,s:\ \exists a,\ c(s,a)\le b,\ \operatorname{Succ}(s,a)\subseteq X\,\}.
\]
If $s\in W_b$, the fixed-point condition provides an admissible action whose successors remain in the same set. We can therefore apply the condition at every subsequent step. Conversely, suppose a state is removed during the fixed-point iteration. Every action either exceeds $b$ immediately or has a possible successor that was removed earlier. Induction on the removal index then constructs a continuation on which harm exceeds $b$.

The composition statement follows by applying the definition of $A_*$ at each step, independently of reward. For targets in $\{0,\tfrac15,1\}$, an error below $\tfrac1{10}$ preserves the strict comparisons required by the veto. The veto takes actions at or above $\tfrac12$ when such actions exist, and otherwise takes a largest value. In either case, the selected action remains in $A_*$. Each accepted update satisfies the same strict error bound, so induction over the accepted updates preserves this conclusion and the preceding bounds on every continuation.

\subsection{Proof of \texorpdfstring{Theorem~\ref{thm:family-transfer}}{Theorem 5} (Transfer across the world family)}\label{proof:thm:family-transfer}

\emph{Step 1, factorisation of the complete recursion.}
Let $L_Nq=q\circ\rho_N$ and let $T_N$ be the right-hand-side operator
of~\eqref{eq:transfer-concrete}. Fix a table $q$. By (C1), the immediate-grade term depends only on the role. For the successor term, equation~\eqref{eq:transfer-menu} and (C2) give
\[
 \min_{s'\in\operatorname{Succ}_N(s,a)}\max_{b\in A_N(s')}q(\rho_N(s',b))
 =\min_{B\in\mathcal K(\rho_N(s,a))}\max_{u\in B}q(u).
\]
Hence $T_NL_N=L_N\mathcal F$. Starting at
$G_{N,0}=\ell_N=L_Nq_0$, induction on the iteration index gives
$G_{N,j}=L_Nq_j$ for every $j$, where $q_{j+1}=\mathcal Fq_j$.
Since the operator is monotone and $q_1\le q_0$, the sequence is decreasing. Its entries remain in the three-element chain $L$, so only finitely many strict decreases are possible before the sequence stabilises. To see that the limit is the greatest fixed point, take any fixed point $u$ of the concrete recursion. It satisfies $u\le\ell_N$, and induction gives $u\le G_{N,j}$ for every $j$. It is therefore bounded above by the limiting concrete table. The same argument applies to the role recursion. This proves
\eqref{eq:transfer-factor} for all worlds, including those with cycles.

\emph{Step 2, why the water family satisfies the conditions for all $N$.}
The state role is $(d,w)$, where $d\in\{0,1,2\}$ and $w$ records water
quality. At $d=0$ there are four action roles: giving, own task, fast
task and swapping. With a request there are five: giving to the
requester, giving to another, and the three non-giving actions. At
$(d,w)=(1,1)$ the future-equivalence state quotient no longer distinguishes the requester, since no continuation can serve the request safely. Giving to the requester and giving to another person therefore form one role. With two steps left, water replacement followed by service still distinguishes the requester, so those giving roles remain separate. Thus there are $4+4+5+4+5+5=27$ roles.
Every role is represented for every $N\ge2$, since the presence of a second person makes giving to someone other than the requester available. The one-step grade depends only on the listed roles, which proves (C1).

For (C2), put $w'=0$ after a swap and $w'=w$ otherwise. The transition rule of the world gives the following complete successor-state-role rule, with the first applicable case taking precedence:
\begin{enumerate}
\item If immediate harm is positive, or the proposed giving action serves
the requester, the sole successor state role is $(0,w')$.
\item Otherwise, if $d=0$, let $U(0)=\{0,1\}$ and $U(1)=\{1\}$.
The successor state roles are $(0,u)$ and $(2,u)$ for $u\in U(w')$.
\item Otherwise the successor state role is $(1,w')$.
\end{enumerate}
Names, channels and noise copies do not change this rule. The menu at
each successor state role consists of the four or five roles just
listed. Consequently the nested families $\mathcal K(r)$, including
the grouping needed for min--max, are independent of $N$. This proves
(C2) on all 27 roles, including the merged urgent poisoned-giving role.

The following deadline calculation verifies the quotient fixed point.
Empty situations admit zero harm by waiting for a request, then, if
necessary, swapping and serving it within its two steps. Thus their
best closed grade is $1$ for either water quality. With $d=1$, clean
water can be given to the requester, giving best grade $1$, whereas poisoned
water allows only lapse or poisoning, giving best grade $1/5$.
With $d=2$, clean water can be served immediately, and poisoned water
can be swapped before serving, giving best grade $1$ in both cases.
More directly, applying $\mathcal F$ to $\bar\ell$ lowers exactly the
own-task role at $(2,1)$ from $1$ to $1/5$: its successor is the
poisoned one-step situation. Every other value stays unchanged, and
the resulting table is a fixed point. Step~1 therefore identifies it
as the greatest fixed point.

The same six menus occur at every $N$. The only non-viable state role
is $(1,1)$, with grade set $\{1/5,0\}$ and no grade-$1$ action.
This proves (C3). There are $8N+2$ states and $2N$ non-viable states,
hence $6N+2$ viable ones. The sole changed role occurs once for every
requester and channel, giving $2N$ changed pairs.

\emph{Step 3, transfer of the certificate.}
Role coverage makes the certified values a complete table on $\mathsf R$. By
Step~1 the reference grades attached to each role are the same in every
world. By (C3), the same role menus occur, so every required strict
comparison in a new non-viable state is already among the certified
comparisons. For a viable state, the weak upper and strict lower
threshold inequalities of Proposition~\ref{prop:decision-region} admit precisely the grade-$1$
actions, which form $A_{*,N}(s)$. In a non-viable state there is no
grade-$1$ action, so the strict lower inequalities exclude every action
at the threshold, while the strict order inequalities place every grade-$1/5$
action above every grade-$0$ action. The fallback maximisers are
nonempty and contained in $A_{*,N}(s)$. This proves (ii). It also shows
why (C3) can be weakened to coverage of just the relevant co-occurring
order pairs, whereas full common-menu coverage is a simple sufficient version.

\emph{Step 4, obstructions and minimal refinement.}
The first assertion of (iii) follows by contraposition from Step~1. To see why the converse fails, observe that the same unavoidable future grade can cap two different immediate labels. Different successor menus can also have the same best closed value. For example, with one action at each
state, let $x,y$ have immediate grades $1,1/5$ and both lead to $z$,
whose grade is $0$ and whose only successor is itself. Giving $x,y$
one role violates (C1), yet both closed grades are $0$. Alternatively,
give all of $x,y,p,q$ grade $1$, let $x\to p$, $y\to q$, with $p,q$
self-looping and assigned distinct roles. Merging $x,y$ violates (C2),
yet their closed grades are both $1$. Thus failure of a strong structural
condition is not itself evidence of an incorrect target.

If $\rho(x)=\rho(y)$ but $g^*(x)\ne g^*(y)$, a scalar function of
that role cannot equal both values. Any partition refining the old
roles on which $g^*$ is constant must separate these two pairs. Every cell of such a partition must therefore lie within a cell of~\eqref{eq:transfer-refinement}. Conversely, the cells defined there refine the old roles and have constant target. They thus form an admissible refinement, and every other admissible refinement is finer. This proves both existence and coarseness.
This paper's two transfer requirements can thus be read as
preserving a suitable role-level law and avoiding broken identifications.
The raw nested-set condition is a sufficient structural certificate,
not a necessary characterisation of every law that happens to preserve
one fixed target.\hfill$\square$

\subsection{State classes and pair roles are different objects}\label{app:state-classes}

Let $C_N$ be the future-state quotient computed by partition refinement with named actions fixed. Its classes are two empty classes,
$N$ clean two-step classes, $N$ clean one-step classes, $N$ poisoned
two-step classes, and one urgent poisoned class: $|C_N|=3N+3$.
The indicated classes are stable under the successor rule, and
different requesters are distinguished immediately when urgent and
clean, after an own step when clean with two steps, or after swapping
when poisoned with two steps. Water and deadline distinguish the other
classes. This also proves that the stable partition has exactly the
listed classes.

A state class is not literally contained in a role fibre, since the
domains are $S_N$ and $X_N$. On the action-marked lift
$(s,a)\mapsto([s],a)\in C_N\times A_N$, each fibre is contained in
one role fibre. The 27 roles are precisely the orbits of this lifted
set under simultaneous person renaming of state and action. The six
state orbits have stabiliser-fixed dimensions $4,4,4,5,5,5$.
The finite-group dimension identity gives these dimensions. To establish that the chosen dynamics or target factors through the roles, we also need (C1)--(C2).

If instead a pair is identified solely by its immediate grade and the
set of successor future-state classes, that pair equivalence and the
27-role partition are incomparable in the checked worlds. For example,
with no request and clean water, own work and fast work have the same
outcome and successors but different roles. Conversely, own work with
clean water and two steps has the same role for two different requesters,
but leads to different urgent requester classes. The closing test instead labels each pair by its future-state class and its step role. Those labels give 73 groups in the full world and 19 on its upbringing domain. They retain the distinction between giving to the requester and giving to another person even in the urgent poisoned state class, and are therefore distinct from the 27 shared roles.

\subsection{Proof of \texorpdfstring{Theorem~\ref{thm:fixed-preservation}}{Theorem 6} (Preservation under continued learning)}\label{proof:thm:fixed-preservation}

We write $\ell(s,a)\in\{0,\tfrac15,1\}$ for the one-step grade and define the closure in grade form,
\[
g^{*}(s,a)\;=\;\min\Bigl(\ell(s,a),\ \min_{s'\in\operatorname{Succ}(s,a)}\ \max_{a'} g^{*}(s',a')\Bigr),
\]
the largest solution, reached by iterating from $g=\ell$ downward. For the integer-arithmetic checks, rescale the harm grades as $\widehat c=5c$ and $\widehat C=5C=5(1-g^*)$, with values $0,4,5$. The equivalent cost form is $\widehat C(s,a)=\min\{b\in\{0,4,5\}:\ \widehat c(s,a)\le b,\ \operatorname{Succ}(s,a)\subseteq W_{b/5}\}$, where $W_{b/5}$ uses the normalised scale defined above. We checked the two forms agree on all $238$ pairs.

\emph{Step 1, the role quotient.} Theorem~\ref{thm:family-transfer} supplies the shared target and certificate coverage. Its proof establishes (C1)--(C3) for every $N\ge2$: the quotient has six state menus and 27 pair roles, its only changed immediate grade is the own task at $(d,w)=(2,1)$, and its only non-viable state menu is $(1,1)$ with values $\{1/5,0\}$. Thus non-viable situations contain no $g^*=1$ action. The exact checks at $N=2,\ldots,6$ supplement this symbolic argument.

\emph{Step 2, contraction.} Fix the target $y=g^{*}$ on the roles. Subtracting the target from the update gives $v_{t+1}-g^{*}=(1-\alpha)(v_t-g^{*})$ in each coordinate. Repeating this identity yields $\sup_r|v_t-g^{*}|=(1-\alpha)^{t-t_0}\sup_r|v_{t_0}-g^{*}|$. By hypothesis, no other gradient reaches $v$, so this recurrence describes the entire change in the table. The same table is used in every world of the family, and the bound therefore applies to all of them. This proves (a).

\emph{Step 3, the two margins.} In a viable situation, some action has $g^{*}=1$, while actions outside this group have $g^{*}\le\tfrac15$. Suppose $\sup|v-g^{*}|<\tfrac3{10}$. Then $v>\tfrac7{10}$ for the first group and $v<\tfrac12$ for the second, giving $\{a:\ v(s,a)\ge\tfrac12\}=\{a:\ g^{*}(s,a)=1\}=A_*(s)$. This set is nonempty, so the veto executes within it whatever the policy proposes. By Step~1, every action in $A_*(s)$ keeps the recipient in $W_0$. The executed action therefore does so as well, proving (b).

In a non-viable situation, all values are at most $\tfrac15$, and both $\tfrac15$ and $0$ occur, as we checked in each of the eight such situations. The veto must then take an action with largest $v$. If $\sup|v-g^{*}|<\tfrac1{10}$, the values of the $\tfrac15$-actions exceed $\tfrac1{10}$, whereas the values of the $0$-actions fall below it. Consequently, every action with the largest $v$ is a $\tfrac15$-action and hence lies in $A_*(s)$, proving (c). The smaller margin is needed here. For example, $v=(0.01,\,0.29)$ on a $(\tfrac15,0)$ pair has both errors below $\tfrac3{10}$, yet the order is reversed.

\emph{Step 4, mixing.} Let $\phi(x)=(1-f)x+f\lambda$ with $0\le f<1$. Since $\phi$ is increasing, it satisfies $\phi(\min X)=\min\phi(X)$ and $\phi(\max X)=\max\phi(X)$ for every finite $X$. We may therefore apply $\phi$ to both sides of the defining recursion. This shows that $\phi\circ g^{*}$ solves the recursion with grade $\phi\circ\ell$.

To identify the greatest solution, compare the two iterations from their initial grades. Iteration from the mixed grades is the image under $\phi$ of iteration from $\ell$ at every step. Both iterations stabilise on a finite set of grade levels and give the greatest solutions, so the same relation holds at their limits. The closure of the mixed grades is therefore $(1-f)g^{*}+f\lambda$. This is also the fixed point of the contraction with mixed targets. We checked the relation with exact fractions for five $(f,\lambda)$ pairs on all $238$ pairs.

\emph{Step 5, decisions under mixing.} Apply Proposition~\ref{prop:decision-region} with $\lambda_r\equiv1$. Its (U), (L) and (O) inequalities are exactly the calibrated conditions used here. The mixed target lies in $\mathcal R$ exactly for $f<3/8$, and simultaneous equal-rate updates preserve membership. At $f=3/8$ the lapse slack obeys $1/2-v_{t+1}=(1-\alpha)(1/2-v_t)$, so a fixed rate below one approaches but never reaches this excluded face, while a unit step reaches it. For $f>3/8$, the first update with $v_t\ge1/2$ is the first $t$ for which $(1-\alpha)^t\le(y-1/2)/(y-v_0)$. For $\alpha=1/10$ and $v_0=1/5$ this gives $27,18,14,11$ at $f=0.40,0.45,0.50,0.55$, as in the joint test of Section~\ref{sec:preservation}. The unequal-rate counterexample and the exact endpoint conventions are proved in Proposition~\ref{prop:decision-region}.\hfill$\square$

\subsection{Proof of \texorpdfstring{Proposition~\ref{prop:decision-region}}{Proposition 2} (The region of correct decisions)}\label{proof:prop:decision-region}

\emph{Step 1, geometry and decisions.}
The box is an intersection of weak affine half-spaces, to which (U) adds further weak half-spaces and (L) and (O) add strict half-spaces. Since each of these sets is convex, their intersection is convex as well. At $g^*$, (U) has
slack $1/2$, (L) has slack $3/10$ or $1/2$, and every (O) difference
is $1/5$, proving nonemptiness and membership of $g^*$.
For $s\in W_0$, (U) includes exactly its $g^*=1$ actions and (L)
excludes all other actions. This threshold set is nonempty and is
$A_*(s)$. For $s\notin W_0$, (L) excludes every action at the
threshold, while (O) places every $1/5$ action strictly above every $0$
action. The nonempty fallback maximisers therefore form a subset of
$A_*(s)$. Nothing in (O) requires distinct $1/5$ actions to remain
tied.

To see the topological qualification, give every role in $I_+$ value
$\tau$, every $1/5$ role value $1/4$, and every $0$ role value $0$.
This table belongs to $\mathcal R$ but arbitrarily small decreases of
a coordinate in $I_+$ violate (U), even within the box. Thus
$\mathcal R$ is not relatively open. Conversely, tables assigning
$1$ to $I_+$, $\tau-1/n$ to the $1/5$ roles and $0$ to the $0$
roles belong to $\mathcal R$ for $n\ge3$, but their limit violates
(L). Thus it is not relatively closed either.

\emph{Step 2, preservation and synchrony.}
Use affine slacks
\[
 q^+_r(v)=v(r)-\tau,\quad
 q^-_r(v)=\tau-v(r),\quad
 q^o_{rr'}(v)=v(r)-v(r').
\]
For a common rate $\alpha$, each satisfies
\[
 q((1-\alpha)v+\alpha y)
 =(1-\alpha)q(v)+\alpha q(y).
\]
If the endpoint slacks are nonnegative, this identity preserves (U) and the box. If they are positive, it also preserves (L) and (O), including when $\alpha=1$. Induction over the updates proves (ii).

For the boundary case, suppose both endpoint slacks are nonnegative. Their combination is strictly positive if and only if at least one positive slack has positive weight. Thus a positive current slack and a zero target slack suffice when $\alpha<1$, but not when $\alpha=1$. We use this distinction in (iv).

Separate coordinate rates keep every coordinate between its two
endpoint values. This preserves the box and the one-coordinate
families (U) and (L), but not differences of coordinates in (O).
For a concrete counterexample on the present 27 roles, let $v$ assign
$1$ to $I_+$, $21/100$ to every $1/5$ role and $0$ to every $0$
role. Let $y=(3/4)g^*+(1/4)\mathbf1$, with values
$(1,2/5,1/4)$ on the three grade levels. Both tables belong to
$\mathcal R$. Update all $1/5$ roles at rate $1/100$, all $0$
roles at rate $1$, and $I_+$ at rate $1/2$. The two lower values
become
\[
 \frac{99}{100}\frac{21}{100}
 +\frac1{100}\frac25=\frac{2119}{10000}
 <\frac14.
\]
The update therefore violates (O), even though every coordinate remains between its endpoint values and (U) and (L) still hold. Separate coordinate rates cannot in general preserve the region. If, however, the rates agree across each edge of $E$, we can apply the common-rate slack identity on that edge. This proves the stated sufficient condition.

\emph{Step 3, necessary and sufficient mixture inequalities.}
The box is automatic because $g^*,\Lambda\in[0,1]^{27}$ and
$0\le f\le1$. Substituting~\eqref{eq:region-mixed} into (U)
gives $g^*(r)+f(\lambda_r-g^*(r))\ge\tau$. If its slope is
negative, rearrangement gives~\eqref{eq:region-upper}, with a weak
endpoint, but otherwise it remains satisfied. Substitution into (L)
gives $g^*(r)+f(\lambda_r-g^*(r))<\tau$. A positive slope yields
\eqref{eq:region-lower}, with a strict endpoint, whereas a nonpositive slope
cannot violate it. Subtracting the two mixed values in (O) gives
\eqref{eq:region-order}. If $\lambda_{r'}>\lambda_r$, solving
this strict inequality gives~\eqref{eq:region-order-critical}.
Equal labels give $(1-f)\Delta_{rr'}>0$, hence $f<1$, whereas labels in
the correct strict order give a convex combination of two positive
differences, positive even at $f=1$. These exhaust the defining
inequalities and prove both necessity and sufficiency. Intersecting
their admissible intervals with $[0,1]$ gives the exact feasible
mixture interval, retaining each endpoint's strictness.

For common $\lambda$, the difference is
$(1-f)(g^*(r)-g^*(r'))$. In particular, for $\lambda=1$,
substitution into (L) gives
\[
 \frac{\tfrac12-\tfrac15}{1-\tfrac15}=\frac38,
 \qquad
 \frac{\tfrac12-0}{1-0}=\frac12.
\]
At the first value the lapse target equals $1/2$ while the poisoning
target is $3/8$: the strict order is still correct. At the second
value the targets are $(1,3/5,1/2)$, again strictly ordered. The
failure is admission at the threshold, not a reversal of order.

\emph{Step 4, boundary targets.}
To identify the closure, replace the strict inequalities by weak ones. Continuity shows that every point in the closure satisfies these weak inequalities. Conversely, take a point satisfying them all and mix it with $g^*$ at positive weight. The resulting point has positive slack in every strict inequality and therefore belongs to $\mathcal R$. As the weight decreases to zero, these points approach the chosen point, proving the reverse inclusion.
For a fixed boundary target with $q(y)=0$, the slack identity in
Step~2 becomes $q(v_{t+1})=(1-\alpha_t)q(v_t)$, which iterates
to~\eqref{eq:region-boundary-product}. A finite product of positive factors is positive. The product becomes zero exactly when a unit rate first occurs, under the assumed strictly positive starting slack.
When all rates are below $1$, the infinite product is zero if and only if $\sum_t-\log(1-\alpha_t)$ diverges. This is equivalent to divergence of $\sum_t\alpha_t$. In one direction, $-\log(1-x)\ge x$ proves the implication. In the other, if the sum of the rates is finite, the rates eventually satisfy $x\le1/2$, where $-\log(1-x)\le2x$. The logarithmic sum is then finite as well.

The convergence qualification is essential. With
$\alpha_{t_0+k}=1/(k+2)^2$ one has
\[
 \prod_{k=0}^{n-1}\left(1-\frac1{(k+2)^2}\right)
 =\frac{n+2}{2(n+1)}\longrightarrow\frac12,
\]
so a positive starting slack never approaches zero despite every
rate being in $(0,1)$. Finally, a zero slack is excluded by (L)
and (O) but allowed by (U). For example, $\Lambda\equiv0$ gives
feasible mixtures $0\le f\le1/2$: at $f=1/2$ the top grade is
exactly $\tau$, while the lower grades are $1/10$ and $0$.
Every $f>1/2$ fails (U), so $1/2$ is an infimum of failing weights,
not a smallest failing weight. This proves (iv).\hfill$\square$

\subsection{Proof of \texorpdfstring{Proposition~\ref{prop:calibrated-codes}}{Proposition 3} (Feasibility of calibrated codes)}\label{proof:prop:calibrated-codes}

Suppose first that a calibrated readout exists. If (i) failed, the value at the same code would have to be both at least and below $\tfrac12$, which is impossible. If (ii) failed, following the strict inequalities around a cycle would require a value to exceed itself. Both conditions are therefore necessary.

Conversely, assign value $1$ to every code carrying a $g^{*}=1$ pair. Condition (i) ensures that no such code also carries a lower-level pair. The remaining finite graph is acyclic by (ii), so we can order its vertices topologically and assign values in $[0,\tfrac12)$ that strictly decrease along every edge. The resulting readout has the required threshold set in viable states. In non-viable states, its maximisers form a subset of the $\tfrac15$-actions, as required by the guarantee.

In the present world, one code is needed above the threshold and two ordered codes below it. Together with the construction just given, this proves the three-code statement for calibrated comparisons. Finally, consider whether $g=w'\circ\phi'$ exists. It does if and only if every pair of inputs assigned the same new code has the same old $g$ value, since this common value both determines and defines $w'$ on that code. The merge counts in the main text were obtained by exhaustive enumeration. \hfill$\square$

\subsection{Proof of \texorpdfstring{Theorem~\ref{thm:representation-preservation}}{Theorem 7} (Preservation during representation learning)}\label{proof:thm:representation-preservation}

Put $z_i=y_i\phi_\theta(x_i)$ and let $c$ be the point of $\operatorname{conv}\{z_i\}_i$ closest to zero. By definition, $\rho(\theta)=\|c\|$. Suppose first that $c\ne0$. The projection inequality gives $z_i^\top c\ge\|c\|^2$ for every $i$, so the readout $v_*(\theta)=mc/\|c\|^2$ satisfies every margin constraint. Its norm is $m/\rho(\theta)$.

We next show that no feasible readout can have a smaller norm. Since $c=\sum_i\lambda_i z_i$ with $\lambda_i\ge0$ and $\sum_i\lambda_i=1$, every feasible $v$ satisfies
\[
 m\le\sum_i\lambda_i z_i^\top v=c^\top v\le\rho(\theta)\|v\|.
\]
The lower bound agrees with the norm of the constructed readout, giving $\beta_m(\theta)=m/\rho(\theta)$ when $\rho>0$. When $\rho=0$, the same inequality excludes every positive-margin readout. Hence a readout satisfying the norm bound exists if and only if $B\rho\ge m$.

Each branch of the update rule returns a feasible pair. Starting from a feasible pair, induction therefore preserves all certified decisions through any sequence of proposed updates. To extend the conclusion from the certificate to the whole family, take a reachable pair $x=(s,a)$ and choose a representative $x_i$ with $r(x_i)=r(x)$. The fixed-role form gives $\phi_\theta(x)=\phi_\theta(x_i)$ at every accepted parameter value. Membership in $A_*(s)$ is constant on each role, so the certified sign at $x_i$ is also the correct sign at $x$. The neural permitted set consequently equals $A_*(s)$ throughout the family, and Theorem~\ref{thm:future-gate} gives the continuation guarantee. Finally, a feature perturbation $e$ changes the score by at most $B\|e\|$. Its sign is therefore preserved whenever $\|e\|<m/B$.

The rejection test can use any nonnegative convex weights and need not compute the nearest point exactly. If those weights satisfy $B\|\sum_i\lambda_i z_i\|<m$, the same norm bound proves that every readout within the permitted norm fails at least one required margin. At seed 22, step 45, the saved weights give $0.4972360012873831<0.5$. The independent verifier recomputes this inequality from the candidate hidden parameters, including a 60-digit evaluation of the $\tanh$ features. The implication follows algebraically, while this numerical check verifies that the saved witness satisfies its premise. \hfill$\square$

\subsection{Proof of \texorpdfstring{Proposition~\ref{prop:relu-factorisation}}{Proposition 4} (Structural factorisation of a shallow ReLU measure)}\label{proof:prop:relu-factorisation}

First consider (i). If $A$ and every $a_s$ belong to
$\operatorname{row}(P)$, write $A=P^\top u$ and $a_s=P^\top u_s$.
Then the function
\[
 L(q)=b+u^\top q+\sum_s r_s\operatorname{ReLU}(u_s^\top q+\beta_s),
 \qquad q\in P(\Omega),
\]
satisfies $F=L\circ P$. Equivalently, every summand is constant on each
$P$-fibre. This direction does not require any domain invariance.

Conversely, suppose $F=L\circ P$ on $\Omega$. At a point $h$ of an
activation cell and for any $n\in K$, sufficiently small positive and
negative $t$ keep $h+tn$ in that cell, because the cell is open.
Since $P(h+tn)=Ph$, factorisation gives $F(h+tn)=F(h)$. On the cell the function is affine, so this equality implies $J_C^\top n=0$. It follows that $J_C\in K^\perp=\operatorname{row}(P)$ for every nonempty cell. We have used only equality on fibres, and no derivative or continuity assumption on $L$ is needed.

Fix a hinge $s$. The set $H_s\cap\Omega$ is nonempty and relatively open
in $H_s$. For each $t\ne s$, distinctness of the hyperplanes implies
that $H_s\cap H_t$ is empty or a proper affine hyperplane of $H_s$.
The finite union of these intersections has empty relative interior, so its complement is relatively open and dense. We can therefore choose a point
\[
 q\in(H_s\cap\Omega)\setminus\bigcup_{t\ne s}H_t.
\]
When $d=1$, $H_s$ is a single point and is disjoint from every other hinge,
so the same assertion holds. Choose a sufficiently small open ball about
$q$, contained in $\Omega$ and disjoint from all other hinge hyperplanes.
Its two sides relative to $H_s$ belong to cells $C_+$ and $C_-$ whose
sign patterns differ only at $s$. Their gradients obey
\[
 J_{C_+}-J_{C_-}=r_sa_s.
\]
Both gradients lie in the linear subspace $\operatorname{row}(P)$, so their difference lies there as well. Since $r_s\ne0$, we obtain $a_s\in\operatorname{row}(P)$. Apply this argument to every hinge and then choose any nonempty cell. Subtracting the active hinge contributions from its gradient gives $A=J_C-\sum_{s\ {\rm active\ on}\ C}r_sa_s\in\operatorname{row}(P)$. Such a cell exists because finitely many hyperplanes cannot cover a nonempty open set. This proves necessity. In particular, the argument
uses only local shifts inside the domain, so dropping $\Omega+K=\Omega$
does not reduce the criterion to sufficiency alone. Without the full-fibre assumption, the requirement that
each hinge meets the domain is essential: on
$\Omega=(-1,1)\times(-2,-1)$, the function
$F(h_1,h_2)=\operatorname{ReLU}(h_2)=0$ factors through $P(h)=h_1$
although its inactive hinge direction does not lie in $\operatorname{row}(P)$.
For coincident hinges, the identity
$\operatorname{ReLU}(-z)=\operatorname{ReLU}(z)-z$ justifies the orientation
and affine absorption stipulated in the canonical form.

For (ii), the case $\delta_C=0$ is immediate. Otherwise put
$u=J_{\perp,C}/\delta_C$ and $h_\pm=h_0\pm ru$.
The ball hypothesis places both points in $C$, and $u\in K$ gives
$Ph_+=Ph_-=Ph_0$. Since $F$ is affine on $C$ and
$J_{P,C}\perp J_{\perp,C}$,
\[
 F(h_+)-F(h_-)=J_C^\top(2ru)=2r\delta_C.
\]
The common value $\ell=L(Ph_0)$ therefore satisfies
\[
 |F(h_+)-\ell|+|F(h_-)-\ell|
 \ \ge\ |F(h_+)-F(h_-)|=2r\delta_C.
\]
The sum of the two errors has the stated lower bound, so at least one error is at least $r\delta_C$. This proves the bound on the ball. If every nonempty cell contains a ball of the same radius, the supremum on $\Omega$ is at least each of the corresponding ball suprema. It is therefore at least $r\max_C\delta_C$, where the maximum exists because there are only finitely many cells.
\hfill$\square$

\subsection{The factorisation audit}\label{app:factorisation-audit}

The ambient domain here is $\mathbb R^{22}$, not the discrete set of valid
world inputs. For an input $h_i$ lying in a cell $C_i$, put
\[
 J_i=A+\sum_{s\ {\rm active\ at}\ h_i}r_sa_s,
 \qquad
 r_i=\min_s\frac{|a_s^\top h_i+\beta_s|}{\|a_s\|},
 \qquad \delta_i=\|NJ_i\|,
\]
where $N$ is the orthogonal projector onto $K$.
By construction, $r_i$ is the distance to the nearest hinge hyperplane. Every smaller closed ball lies in $C_i$, whereas $B(h_i,r_i)$ may touch the boundary. Since $F$ is continuous, its affine formula on the cell extends to the cell closure. We can therefore apply the endpoint argument of Proposition~\ref{prop:relu-factorisation}(ii) to this limiting ball as well. If $r_i=0$ or $\delta_i=0$, the corresponding lower bound is zero.
Otherwise the endpoints $h_i\pm r_iNJ_i/\delta_i$ have the same
observation $Ph_i$ and logits $F_i\pm r_i\delta_i$, where $F_i=F(h_i)$.
For any reduced gate map $G$, the triangle inequality therefore gives
\[
 \sup_{h\in B(h_i,r_i)}|g(h)-G(Ph)|
 \ \ge\ \frac12\left|
 \sigma(F_i+r_i\delta_i)-\sigma(F_i-r_i\delta_i)
 \right|.
\]
Taking the maximum over the sampled inputs yields $B_g$ on their union of
balls. The logit-scale diagnostic instead takes $\max_i r_i\delta_i$, and
neither calculation assumes a common radius for all cells or enumerates
every cell. The reported $0.0084$--$0.0902$ values are the gate-scale
diagnostic, not this logit bound and not the local approximation
$r_i\sigma'(F_i)\delta_i$.
The audit implements the observation through the orthogonal projection
$I-N$, and a full-row-rank matrix whose rows form an orthonormal basis of
$K^\perp$ has the same fibres and satisfies $P^\top P=I-N$.
Finally, if $g_\theta(h)=\Phi_\theta(Ph)$ for every $\theta$, then
$g_\theta(h)=g_\theta(h')$ whenever $Ph=Ph'$, for every $\theta$.

\end{document}